%% file: main.tex
\documentclass[10pt,twocolumn,letterpaper]{article}

\usepackage{wacv}              

\input{preamble}

\definecolor{wacvblue}{rgb}{0.21,0.49,0.74}
\usepackage[pagebackref,breaklinks,colorlinks,allcolors=wacvblue]{hyperref}
\usepackage{array}
\usepackage[accsupp]{axessibility}
\newcommand{\name}{MedHiVE}
\newcommand{\layername}{HiVEL}
\def\wacvPaperID{155} 
\def\confName{WACV}
\def\confYear{2027}

\title{Multimodal Hierarchical Mixture-of-Experts \\ for 3D Medical Vision-Language Modeling}

\newcommand{\midsize}{\fontsize{10.2pt}{11pt}\selectfont}

\author{
\makebox[\textwidth][c]{%
Gizem Mert$^{1,2,3}$ \thanks{Corresponding author: \texttt{gizem.mert@helmholtz-munich.de}} \quad
Szymon Płotka$^{4}$ \quad
Pedro R. A. S. Bassi$^{5}$ \quad
Wenxuan Li$^{5}$%
} \\
\makebox[\textwidth][c]{%
Zongwei Zhou$^{5}$ \quad
Jarosław B. Ćwikła$^{6}$ \quad
Arkadiusz Sitek$^{7,8}$ \quad
Ewa Szczurek$^{1,3,9,10}$%
} \\[0.3em]
\makebox[\textwidth][c]{%
\midsize$^{1}$ Helmholtz Munich \quad
$^{2}$ Technical University of Munich \quad
$^{3}$ Munich Center for Machine Learning%
} \\
\makebox[\textwidth][c]{%
\midsize$^{4}$ Jagiellonian University \quad
$^{5}$ Johns Hopkins University \quad
$^{6}$ GAMMED \quad
$^{7}$ Massachusetts General Hospital%
} \\
\makebox[\textwidth][c]{%
\midsize$^{8}$ Harvard Medical School \quad
$^{9}$ University of Warsaw \quad
$^{10}$ LMU Munich%
}
}

\begin{document}
\maketitle
\input{sec/0_abstract}    
\input{sec/1_intro}
\input{sec/4_related_work}
\input{sec/Preliminaries}
\input{sec/5_methodology}
\input{sec/6_experiments}
\input{sec/7_conclusion}
\input{sec/acknowledgments}

{
    \small
    \bibliographystyle{ieeenat_fullname}
    \bibliography{main}
}

\input{sec/appendix}

\end{document}

%% file: preamble.tex
\usepackage{tikz}
\usetikzlibrary{arrows.meta}
\definecolor{qwenpurple}{RGB}{97, 92, 237}

\usepackage{pifont}
\newcommand{\cmark}{\ding{51}}
\newcommand{\xmark}{\ding{55}}
\usepackage{cuted}
\usepackage{tcolorbox}
\tcbuselibrary{skins,breakable}
\usepackage{xcolor}
\definecolor{reportblue}{RGB}{33, 64, 154}
\definecolor{jsonteal}{RGB}{54, 117, 136}
\usepackage{amsmath}
\usepackage[T1]{fontenc}


%% file: sec/0_abstract.tex
\begin{abstract}
  Medical 3D scans and reports are inherently hierarchical, with diagnostic cues concentrated in specific anatomical regions across multiple levels of organization. Yet, most medical vision-language models ignore this structure and fail to focus on clinically relevant anatomy. We introduce \name{}, which combines hierarchical vision-language Soft Mixture-of-Experts (SMoE) with contrastive learning to jointly model fine-grained anatomy and global scan context. 
  To our knowledge, \name{} is the first framework to route both vision and text tokens through a two-level SMoE hierarchy organized by anatomical scale, coupling this with report-derived anatomy-level contrastive alignment for 3D medical imaging. This replaces the direct region–text alignment of statically pooled features and vision-only hierarchical routing of prior work with dynamic, slot-based cross-modal information flow across anatomical scales. \name{} operates via a two-level, learnable token-routing mechanism, where local image and text features are first routed through a local SMoE and then fused by a global SMoE into unified multimodal representations. Contrastive learning aligns both anatomy-level and global vision-language representations in a unified space. Pre-trained on 25,000+ CT scans with anatomical supervision, \name{} consistently surpasses 11 state-of-the-art methods across 6 datasets on 3 tasks: segmentation, classification, and detection, demonstrating the importance of hierarchical modeling. The code for \name{} is at \href{https://github.com/szczurek-lab/MedHiVE}{github.com/szczurek-lab/MedHiVE}.
\end{abstract} 

%% file: sec/1_intro.tex
\section{Introduction}
\label{sec:intro}

Three-dimensional (3D) computed tomography (CT) data possess an intrinsic, multi-scale hierarchical structure where local voxel-level textures are nested within anatomical organs, which in turn follow a natural spatial arrangement within the body. Clinical text mirrors this hierarchy, as reports are typically organized into organ-specific findings and global impressions. Unifying these domains requires learning joint representations that bridge fine-grained anatomical details with holistic bodily context, a balance that remains a key challenge for medical vision-language models. 

\begin{figure}[t]
    \centering
    \includegraphics[width=\linewidth]{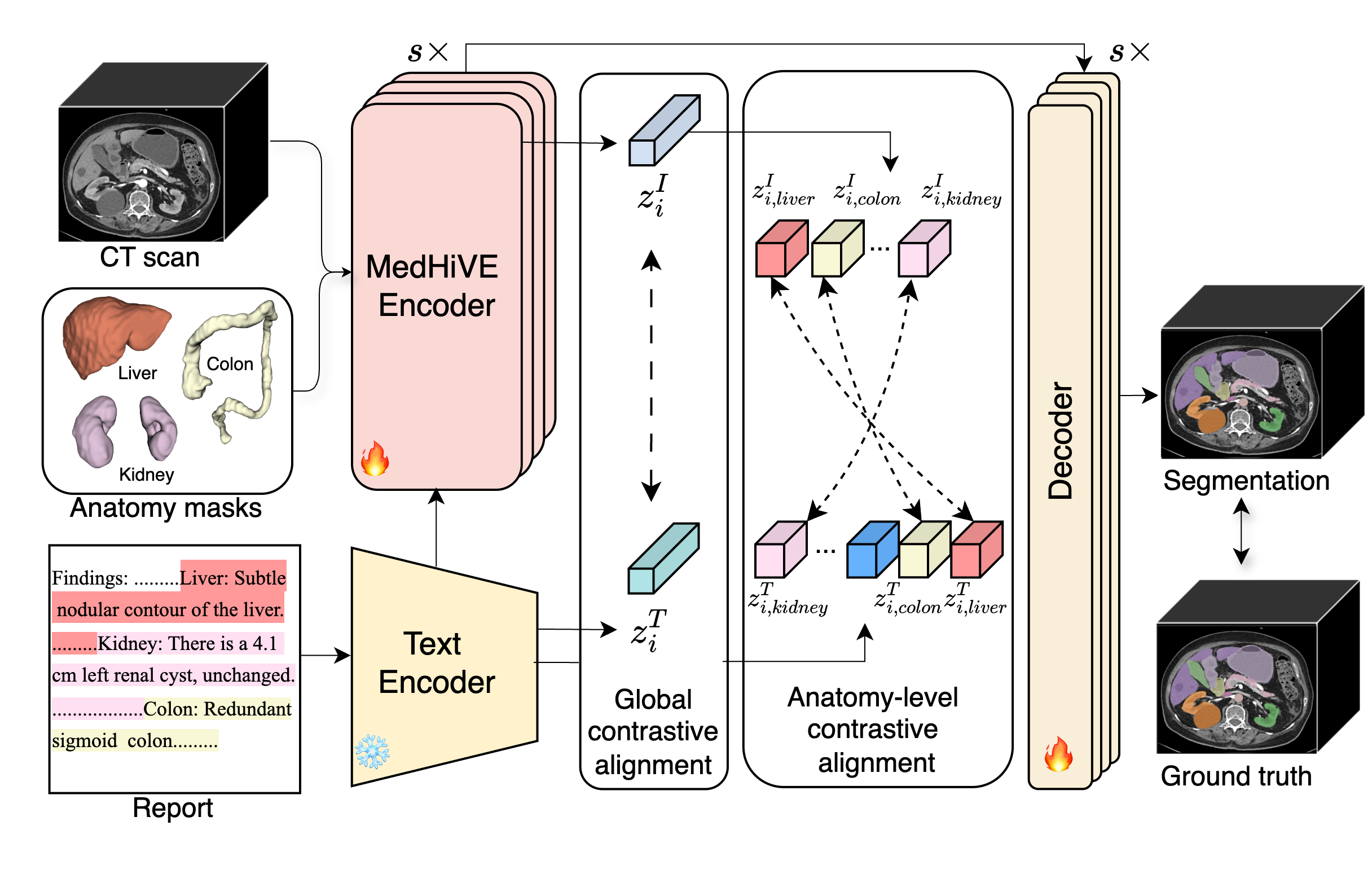}
    \caption{
        Overview of the \name{} framework. A hierarchical encoder processes a 3D CT scan and anatomy masks alongside a full report representation $z_i^{T}$ and anatomy-specific snippet representations $z_{i,a}^{T}$, extracted through a frozen text encoder. The global vision representation $z_i^{I}$ is contrasted with $z_i^{T}$, while local vision representations $z_{i,a}^{I}$ are aligned with $z_{i,a}^{T}$ for each anatomical structure $a$. This yields anatomy-aware vision-language representations that enhance downstream segmentation, classification, and detection.
    }
    \label{fig:placeholder}
\end{figure}

While recent progress in vision-language (VL) models has revolutionized  natural image understanding,   standard 3D medical approaches often treat volumes and reports as flat sequences, effectively discarding these essential structural inductive biases~\cite{hsu2021capturing}. The MambaHoME architecture~\cite{plotka2025mamba} recently introduced hierarchical routing, but solely for vision inputs, and the challenge of jointly modeling vision-language cross-modal hierarchies in a unified representation remains an open problem. Existing medical VL models have largely focused either on global alignment~\cite{boecking2022making} or localized patches~\cite{zhang2022contrastive}. More recent efforts introduce anatomy-aware priors~\cite{yan2024ahive, shui2025largescale, cao2025boosting}, yet they frequently treat image and text hierarchies as separate or auxiliary concerns. To the best of our knowledge, no previous work has unified the hierarchical routing of 3D vision tokens with the semantic hierarchy of clinical reports into a single, differentiable framework for image analysis.

To bridge this gap, we introduce \name, 
a 
\textbf{Med}ical 
\textbf{Hi}erarchical  
\textbf{V}ision--Language
{\textbf{E}}xperts framework, designed to exploit multi-scale structural priors.
We encode each CT volume and its report into global and anatomy-level representations, using a long-context text encoder~\cite{li2023comparative} for the report and voxel-wise organ masks for the volume. We propose a hierarchical multimodal encoder with two-level, slot-based Soft MoE routing over global tokens and anatomy-level representations from both vision and text. 
Global vision tokens and paired global text tokens are softly grouped and processed by global experts, while anatomy-level image and text representations are handled in parallel by local experts. These global and anatomy-aware slots are concatenated and routed through a second SMoE layer with shared experts for joint vision--language processing. 
To enrich learned representations, we apply a contrastive loss to align vision-language representations on both local and global scales.  Our main contributions are as follows:
\begin{enumerate}
\item \textbf{Hierarchical vision--language architecture} operating on anatomy-level representations and global tokens from both vision and text, using slot-based SMoE routing through \emph{anatomy-specific and global vision and language experts}.

\item \textbf{Local-global vision--language contrastive learning} that jointly optimizes scan--report and anatomy--snippet objectives, {\textit{aligning representations across global and anatomical scales}}.

\item \textbf{Anatomy-aware supervision} that combines organ masks with automatically extracted organ-specific report snippets, enabling organ-level alignment \emph{without sentence-level labels}.

\end{enumerate}

We pre-train \name{} on a large-scale, publicly available dataset comprising over 25,000 CT scans with reports and demonstrate consistent gains over state-of-the-art in downstream tasks like segmentation, classification, and detection. Beyond improving segmentation accuracy, \name{} provides a mechanism for transferring knowledge from free-form reports into 3D representations.

%% file: sec/4_related_work.tex
\section{Related work}
\label{sec:related_work}

\noindent
\textbf{Medical Vision-Language Models (MedVLMs).} The emergence of VLMs has catalyzed a shift toward multimodal supervised learning, mapping heterogeneous data into a unified latent space~\cite{radford2021learning, zhang2025multimodal,li2026cancerverse}. In the medical domain, research has evolved from 2D radiograph alignment~\cite{huang2021gloria} to complex 3D volumetric analysis. However, 3D VLMs remain non-trivial due to the semantic density bottleneck~\cite{cao2025boosting}. Recent state-of-the-art 3D MedVLMs address this challenge by improving the quality and scalability of visual representations~\cite{blankemeier2026merlin, hamamci2025better, cao2025boosting}. They enhance the signal-to-noise ratio of visual features; for instance,~\cite{cao2025boosting} utilize anatomical normality modeling via VQ-VAEs to amplify abnormal signals. fVLM ~\cite{shui2025largescale} is proposed to match anatomical regions with descriptions while calibrating for false negatives. Despite these advances, fVLM aligns each anatomical region through fixed, non-parametric pooling at a single level, which leaves the relationship between global physiological context and discrete anatomical structures unresolved. Anatomical normality modeling~\cite{cao2025boosting} likewise treats each anatomy in isolation, enhancing per-anatomy semantics through a multi-stage training pipeline before report alignment. To bridge this gap, we introduce an end-to-end framework that jointly routes anatomy-level and global tokens through learned experts. By capturing both local anatomical details and global physiological context within a single, unified representation, our approach replaces isolated per-anatomy alignment and multi-stage training with learned fusion of anatomy-level and global tokens, while explicitly resolving multi-scale relationships.
\noindent
\textbf{Mixture-of-Experts (MoE).} MoE~\cite{jacobs1991adaptive} decomposes a task by combining multiple specialized sub-networks (i.e., experts) within a single architecture. Sparsely gated MoE layers~\cite{shazeer2017,riquelme2021scaling,baykal2022theoretical,yang2025tra} enable conditional computation by employing a learned router that activates only a sparse subset of experts per input token. This mechanism allows for a massive increase in model capacity while keeping the computational cost nearly constant. Despite their scaling advantages, sparse routers often introduce optimization challenges such as expert imbalance and training instability. Soft MoE~\cite{puigcerver2024from} addresses these issues through a continuous assignment approach, passing weighted combinations of all tokens to each expert, thereby smoothing the optimization landscape while retaining throughput efficiency. While modern MoE was largely popularized by its success in Transformer-based LLMs~\cite{zeng2026s,li2024cumo}, the architecture has rapidly expanded into multimodal domains~\cite{li2025uni,kim2025mohave}, as well as medical imaging~\cite{wu2025dynamic, plotka2025mamba, tagnamas2026diffmoe}, where expert specialization is particularly suited for learning from complex, high-dimensional biological patterns. In this work, we build upon these advancements,
employing a two-level Soft MoE routing mechanism to hierarchically align global and anatomy-specific features across 3D CT scans and medical reports.

\noindent
\textbf{Contrastive learning in MedVLMs.} Contrastive learning remains the primary objective for aligning medical images with clinical reports. While early methods focused on global alignment~\cite{zhang2022contrastive, huang2021gloria}, the field has recently shifted toward structural alignment to respect the inherent hierarchy of medical data. HiCLIP~\cite{genghiclip} pioneered this by using hierarchy-aware attention to capture nested relationships between multi-level visual features and text for 2D imaging. However, extending hierarchical attention to 3D medical analysis remains non-trivial due to the cubic growth of tokens in volumetric scans. Prior works such as GLoRIA~\cite{huang2021gloria} learn global and region-level vision-language alignment via attention mechanisms in 2D medical images, but do not generalize to dense 3D volumes or provide explicit hierarchical modeling and token-level routing across anatomical scales.
While recent 3D approaches like~\cite{shui2025largescale} and~\cite{cao2025boosting} have scaled pre-training to massive 3D datasets, they often rely on coarse-grained global alignment that may overlook fine-grained anatomical morphology. Our work bridges this gap by extending hierarchical contrastive alignment to 3D and introducing a learnable, token-level routing mechanism for multimodal fusion. Unlike the static hierarchical attention in HiCLIP, we rely on a two-level multimodal SMoE routing paradigm, allowing dynamic, slot-based information flow between holistic scan representations and anatomy-aware tokens while maintaining the scalability required for large-scale 3D pre-training.

%% file: sec/Preliminaries.tex
\section{Preliminaries}
\label{preliminaries}

\textbf{Selective State-Space Models and Mamba.} Selective State-Space Models (SSMs) form the foundation of Mamba~\cite{gu2024mamba}, a selective sequence model that provides a scalable alternative to self-attention for capturing long-range dependencies.  
An SSM processes a 1D continuous-time input signal $x(t) \in \mathbb{R}$ through a latent state $h(t) \in \mathbb{R}^N$ and outputs $y(t) \in \mathbb{R}$:
\begin{equation}
\begin{gathered}
h'(t) = \boldsymbol{\Lambda} h(t) + \boldsymbol{\Gamma} x(t), \quad y(t) = \boldsymbol{\Psi} h(t)
\end{gathered},
\end{equation}
where $\boldsymbol{\Lambda} \in \mathbb{R}^{N \times N}$ governs state evolution, $\boldsymbol{\Gamma} \in \mathbb{R}^{N \times 1}$ injects the input, and $\boldsymbol{\Psi} \in \mathbb{R}^{1 \times N}$ maps the state to output space.  
For discrete-time modeling, the system is discretized with step size $\Delta$:
\begin{equation}
\begin{gathered}
\bar{\mathbf{A}} = \exp(\Delta \boldsymbol{\Lambda}), \quad \bar{\boldsymbol{\Gamma}} = (\Delta\boldsymbol{\Lambda})^{-1}(\exp(\Delta\boldsymbol{\Lambda}) - I)\Delta\boldsymbol{\Gamma} \\
h_t = \bar{\mathbf{A}} h_{t-1} + \bar{\boldsymbol{\Gamma}} x_t, \quad y_t = \boldsymbol{\Psi} h_t
\end{gathered}.
\end{equation}

\noindent
Mamba makes this discretized SSM selective, i.e., the effective state transition and input injection are modulated by the current input through learned gating mechanisms, while retaining linear-time parallel scan. This enables efficient modeling of long 1D sequences extracted from high-resolution 3D CT volumes. In our framework, Mamba layers constitute the backbone of the vision-language encoder, providing scalable global context modeling across 
scales.

\noindent
\textbf{Soft Mixture-of-Experts and Slot-Based Routing.} To balance efficiency and capacity, we employ Soft Mixture-of-Experts (SMoE) layers with slot-based routing (see Figure \ref{fig:fig2}). An SMoE layer consists of \textbf{Experts:} A set of $E$ parameterized feed-forward networks (FFNs); and \textbf{Slots:} A collection of $P$ learnable vectors acting as queries for grouping input tokens before expert processing. Given input tokens $\{u_k\}_{k=1}^{K}$ and slot-representation matrix
$\mathbf{E} \in \mathbb{R}^{P \times d}$ (with $P$ slots), the soft assignment of token $u_k$ to slot $p\in\{1,\ldots, P\}$ is:
\begin{equation}
\mathbf{A}[k,p] = \frac{\exp(\langle u_k, \mathbf{E}[p,:] \rangle)}
{\sum_{p'=1}^{P} \exp(\langle u_k, \mathbf{E}[p',:] \rangle)}.
\end{equation}
\noindent
Each slot aggregates the tokens via a sum weighted by the soft assignment values, and the resulting slot representations are processed by allocated experts.  
This mechanism provides dynamic token grouping, improved parameter efficiency, and effective scaling to long sequences like high-resolution 3D volumes.

\noindent
\textbf{Notation.} We denote a dataset of paired samples as $\mathcal{D} = \{(\mathbf{X}_i^{I}, \mathbf{X}_i^{T})\}_{i=1}^{N}$, 
where $\mathbf{X}_i^{I} \in \mathbb{R}^{D \times H \times W}$ is a 3D CT volume and $\mathbf{X}_i^{T}$ is the corresponding radiology report. The minibatch size is denoted by $B$. Predefined anatomical structures are indexed by $a \in \{1, \ldots, A\}$, where $A$ is the maximal number.  For each sample $i$ and anatomy $a$, binary masks $\mathbf{M}_{i,a} \in \{0,1\}^{D \times H \times W}$ indicate the spatial support of anatomy $a$. At encoder stage $s \in \{1,2,3,4\}$, feature maps are denoted by $\mathbf{F}_i^{(s)} \in \mathbb{R}^{d_s \times D_s \times H_s \times W_s}$, 
with $N_s = D_s H_s W_s$ tokens after flattening. Global and anatomy-level image representations are obtained via pooling and denoted by $z_i^{I(s)} \in \mathbb{R}^{d_s}$ and $z_{i,a}^{I(s)} \in \mathbb{R}^{d_s}$, respectively. On the text side, the full report and anatomy-specific snippets are similarly encoded as $z_i^{T(s)} \in \mathbb{R}^{d_s}$ and $z_{i,a}^{T(s)} \in \mathbb{R}^{d_s}$. In the \layername{} layer, tokens are grouped into $G$ groups of size $K$ such that $N_s = G K$. We denote the number of experts by $E$ and the number of slots per expert by $S$, yielding $P = E \cdot S$ slots. Token-to-slot assignments are represented by soft assignment matrices $\mathbf{A} \in \mathbb{R}^{K \times P}$.

%% file: sec/5_methodology.tex
\section{Methodology}
\label{sec:methodology}
\noindent
\textbf{Problem formulation.} Our primary goal is to learn a model that captures multi-level image-text alignment, ensuring that both global representations $(z_i^{I}, z_i^{T})$ and anatomy-level representations $(z_{i,a}^{I}, z_{i,a}^{T})$ are semantically aligned across modalities. We use “global” to denote scan-level summaries and “local” to denote anatomy-conditioned regions; the same hierarchy is mirrored on the text side via report-level and snippet-level representations.


\subsection{Image-text Data Preprocessing}
\label{sec:dataprocessing}

\noindent
\textbf{Report snippet extraction.} Following~\cite{shui2025largescale}, we decompose each report into anatomy-level 
components: $\mathbf{X}_i^{T} \longmapsto \{\mathbf{X}_{i,a}^{T}\}_{a=1}^{A_i}$,
where $\mathbf{X}_{i,a}^{T}$ is the corresponding textual description, and $A_i$ is the number of anatomical structures present in sample $i$. In our implementation, we consider 14 structures in total.  The anatomy-level text is parsed by Qwen3-30B~\cite{yang2025qwen3} applied to the full radiology report to identify and extract organ-specific snippets (see Appendix \ref{appendix:prompts}).

\noindent
\textbf{Anatomy inventory.} For each sample $i$ and anatomy $a$, we maintain a binary label mask $\mathbf{M}_{i,a} \in \{0,1\}^{D \times H \times W}$,
where $\mathbf{M}_{i,a}[v] = 1$ indicates that voxel $v$ belongs to anatomy $a$ in image $i$ and $\mathbf{M}_{i,a}[v] = 0$ otherwise.  
These masks implement the anatomical inventory (see Table~\ref{tab:organs_list}) and are used to derive both anatomy-level and global image representations as explained in Section~\ref{medhive_encoder}.

\begin{figure}
\includegraphics[width=\linewidth]{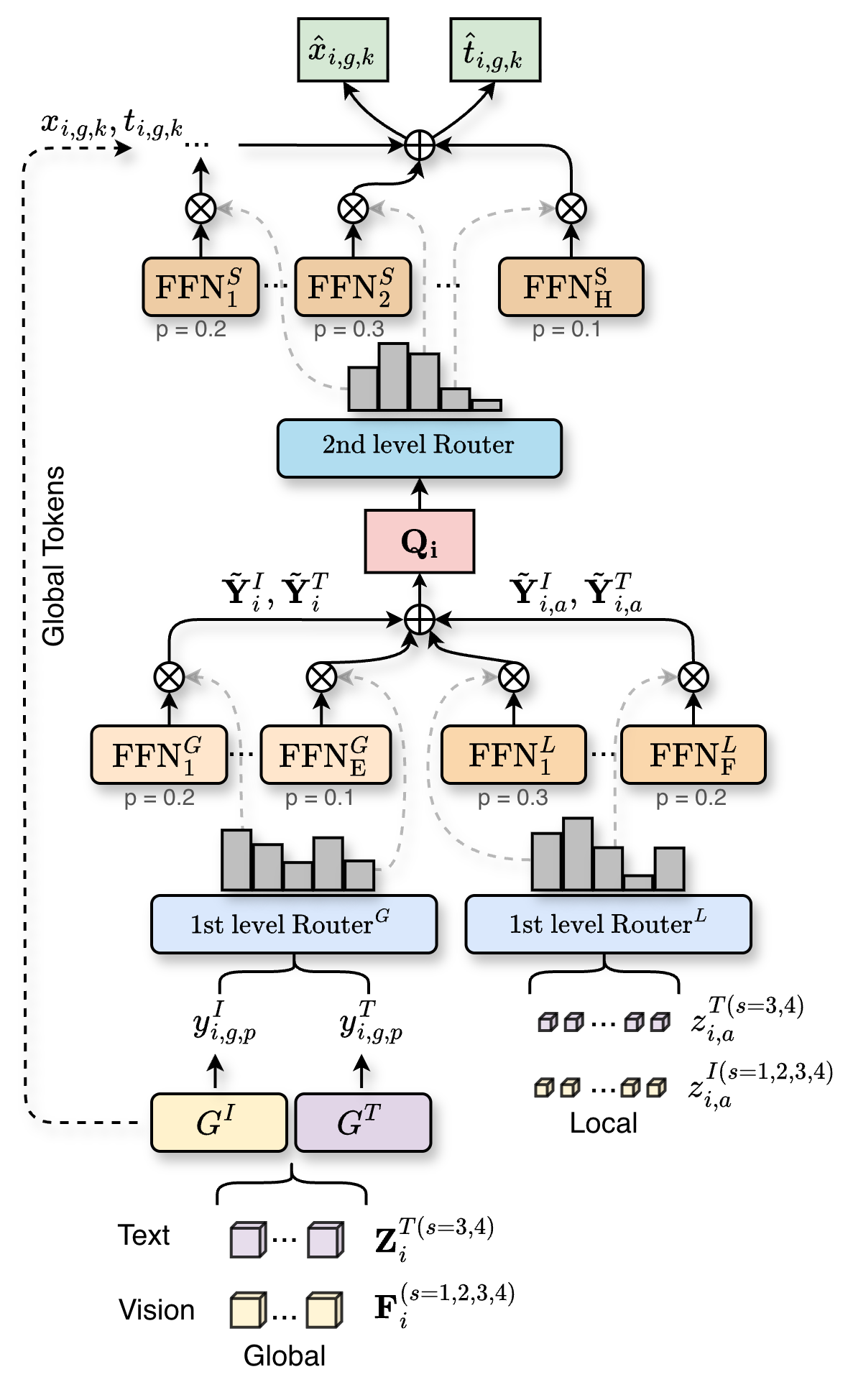}
\caption{
    Overview of the \layername{} layer. Global and anatomy-level vision tokens and text tokens are routed to $E$ global experts $\mathrm{FFN}^{G}$ and $F$ local experts $\mathrm{FFN}^{L}$ via first-level routers. The resulting slots are concatenated and processed by a second-level router. Global vision and text tokens are fused with $H$ shared experts $\mathrm{FFN}^{S}$ outputs via residual connection, yielding enhanced vision tokens $\hat{x}$ and text tokens $\hat{t}$. Vision tokens are used at all stages $s \in \{1,2,3,4\}$, while text tokens are enabled at $s \in \{3,4\}$.
}
\label{fig:fig2}
\end{figure}

\subsection{The \name{} Architecture}
We introduce \name{}, an encoder–decoder architecture featuring a newly designed \name{} Encoder that incorporates \layername{}, a Hierarchical Vision-Language Experts Layer, and applies contrastive learning to achieve multimodal representation alignment.

\subsubsection{Encoder}
\label{medhive_encoder}
We propose a multimodal \name{} Encoder, which extends the vision-only HoME Block of~\cite{plotka2025mamba} with text routing, an anatomy expert stream, second-level cross-modal fusion, and contrastive alignment, illustrated in Figure~\ref{fig:placeholder}. For {\textbf{feature extraction}} at each encoder stage, we follow SegMamba~\cite{xing2024segmamba}, applying 
a 3D Gated Spatial Convolution (GSC) block followed by  flattening the  feature map  into 1D token sequences, normalization, a Mamba layer to capture long-range dependencies, and again normalization.  The feature extraction step is followed by the \layername{} layer (see Figure \ref{fig:fig2}), with additional inputs from text representations and local vision representations, as described below. The output of \layername{} is  flattened and rearranged. Both normalization and Mamba layers, as well as  normalization and \layername{} layers are wrapped in residual connections. 

\noindent
\textbf{Vision representations.} The feature extraction procedure results in global feature maps $\mathbf{F}_i^{(s)} \in \mathbb{R}^{d_s \times D_s \times H_s \times W_s}$  for stages $s=1,2,3,4$, where $d_s$ is the channel dimension and $(D_s, H_s, W_s)$ is the spatial grid at stage $s$. At each stage, the anatomy masks are downsampled to the token grid via a stage-specific adaptive max pooling operator $\mathrm{Down}_s$:
\begin{equation}
\mathbf{M}^{(s)}_{i,a} = \mathrm{Down}_s\big(\mathbf{M}_{i,a}\big) 
\in \{0,1\}^{D_s \times H_s \times W_s}.
\end{equation}
\noindent
Next, anatomy-level image representations are obtained by masked averaging of the tokens:
\begin{equation}
z^{I(s)}_{i,a}
= \mathrm{Pool}\big(\mathbf{F}_i^{(s)} \odot \mathbf{M}^{(s)}_{i,a}\big)
\in \mathbb{R}^{d_s},
\end{equation}
where $\odot$ applies the binary mask and $\mathrm{Pool}(\cdot)$ is mean pooling.
In parallel,  global vision representations are obtained by global pooling:
\begin{equation}
z_i^{I(s)} = \mathrm{Pool}\big(\mathbf{F}_i^{(s)}\big) \in \mathbb{R}^{d_s}.
\end{equation}
Both anatomy-level representations $z^{I(1)}_{i,a}$ and global representations $z_i^{I(1)}$ obtained in stage $s=1$ are used for contrastive alignment, as explained in Section~\ref{training_obj} below. For all stages $s\in\{1,2,3,4\}$, global feature maps $\mathbf{F}_i^{(s)}$ become the input to \layername{} layer, as described in Section~\ref{sec:vlhome}.


\noindent
\textbf{Text representations.}
Each radiology report $\mathbf{X}_i^{T}$ and the corresponding anatomy descriptions $\mathbf{X}_{i,a}^{T}$ are encoded with a frozen Clinical-Longformer~\cite{li2023comparative}, producing representations $\mathbf{T}_i \in \mathbb{R}^{L_i \times d_t}$ for the full report and $\mathbf{T}_{i,a} \in \mathbb{R}^{L_{i,a} \times d_t}$ for each anatomy snippet, where $L_i$ and $L_{i,a}$ are the tokenized lengths of the full report and anatomy text, and $d_t$ is the text hidden dimension. To interface with stage-$s$ visual representations of dimension $d_s$, we project text tokens with a linear map $\mathbf{W}^{(s)} \in \mathbb{R}^{d_s \times d_t}$. For the global text stream, we pad or truncate the report token sequence to the number of visual tokens $N_s=D_sH_sW_s$, yielding $\mathbf{Z}_i^{T(s)} \in \mathbb{R}^{N_s \times d_s}$. For global contrastive alignment (see Section \ref{training_obj}), we obtain the text representation by mean pooling:
\begin{equation}
z_i^{T(s)} = \mathrm{Pool}\big(\mathbf{Z}_i^{T(s)}\big) \in \mathbb{R}^{d_s}.
\end{equation}
For anatomy text, we use a token per anatomy and project it to obtain $z_{i,a}^{T(s)} \in \mathbb{R}^{d_s}$. These global and anatomy-level text tokens are provided to the \layername{} layer (Section~\ref{sec:vlhome}).


\subsubsection{\textbf{\layername{} layer}}
\label{sec:vlhome}

The \layername{} layer takes the 
global vision feature map $\mathbf{F}_i^{(s)}$ as well as local vision representations $z^{I(s)}_{i, a}$ as input, and, for stages $s=3, 4$, processes them on two levels, in conjunction with the text representations $z_{i,a}^{T(s)}$ and $\mathbf{Z}_i^{T(s)}$ (see Figure~\ref{fig:fig2}).

\noindent
\textbf{First level routing.} To perform {\textbf{global routing}}, we first flatten the spatial grid of $\mathbf{F}_i^{(s)}$ into a vision token matrix $\mathbf{V}_i^{I} \in \mathbb{R}^{N_s \times d_s}$ and partition it into $G$ non-overlapping groups of size $K$ where $N_s=GK$. In the following, we skip the stage index $s$ for clarity of notation.

Let $\mathbf{Z}_i^{T} \in \mathbb{R}^{N_s \times d_s}$ denote the corresponding text token matrix, and let $x_{i,g,k},\, t_{i,g,k} \in \mathbb{R}^{d_s}$ be the $k$-th token in group $g$ extracted from $\mathbf{V}_i^{I}$ and $\mathbf{Z}_i^{T}$, respectively. For each group $g$, we compute:
\begin{equation}
\begin{aligned}
\bar x_{i,g} &= \frac{1}{K} \sum_{k=1}^{K} x_{i,g,k}, &
\bar t_{i,g} &= \frac{1}{K} \sum_{k=1}^{K} t_{i,g,k},
\end{aligned}
\end{equation}

\noindent
and use them to generate sample- and group-dependent slots through a projection network shared across groups and modalities. For $P = E_1 \cdot S_1$ slots (with first-level $E_1$ experts and $S_1$ slots per expert), we obtain slot representation matrices $\mathbf{E}_{i,g}^{I}$ and $\mathbf{E}_{i,g}^{T} \in \mathbb{R}^{P\times d_s}$:
\begin{equation}
\begin{aligned}
\mathbf{E}_{i,g}^{I}[p,:] &= \Phi(\bar x_{i,g})_p, &
\mathbf{E}_{i,g}^{T}[p,:] &= \Phi(\bar t_{i,g})_p,
\end{aligned}
\end{equation}
where $\Phi(\cdot)$ maps $\mathbb{R}^{d_s}$ to $P$ slot vectors. We then compute the soft token-to-slot assignment matrices $\mathbf{A}_{i,g}^{I}$ and $\mathbf{A}_{i,g}^{T}$ for vision and text tokens, respectively:
\begin{align}
\mathbf{A}_{i,g}^{I}[k,p] &= \mathrm{softmax}_{p}\big(\langle x_{i,g,k}, \mathbf{E}_{i,g}^{I}[p,:]\rangle\big),\\
\mathbf{A}_{i,g}^{T}[k,p] &= \mathrm{softmax}_{p}\big(\langle t_{i,g,k}, \mathbf{E}_{i,g}^{T}[p,:]\rangle\big).
\end{align}
 Group slot representations are obtained by weighted summing:
\begin{align} y_{i,g,p}^{I} = \sum_{k=1}^{K} \mathbf{A}_{i,g}^{I}[k,p]\, x_{i,g,k}, && y_{i,g,p}^{T} = \sum_{k=1}^{K} \mathbf{A}_{i,g}^{T}[k,p]\, t_{i,g,k} . \end{align}
A lightweight modality router and group gates modulate the relative contribution of vision and text streams, producing per-group, per-modality scaling factors that rescale the slots before expert processing. The resulting slots are passed through $\mathrm{E}$ shared global experts $\mathrm{FFN^G}$:
\begin{equation}
\begin{aligned}
\tilde y_{i,g,p}^{I} &= \mathrm{FFN}^{G}_e\big(y_{i,g,p}^{I}\big),\\
\tilde y_{i,g,p}^{T} &= \mathrm{FFN}^{G}_e\big(y_{i,g,p}^{T}\big),
\end{aligned}
\qquad e \in \{1,\ldots,\mathrm{E}\}.
\end{equation}

We associate a contiguous block of $S_1$ slots with each expert: expert $e$ processes slots $p\in\{(e-1)S_1+1,\ldots,eS_1\}$, and its output is scaled by a learned expert gate. We adopt deterministic block allocation to enforce spatial-semantic locality, ensuring that contiguous anatomical regions and semantically coherent text snippets are routed together. This structural prior is designed to promote training stability and mitigate expert imbalance, while preserving learnable routing via soft assignments. The slots are reshaped into sequences of $P$ slots per group:
\begin{equation}
\begin{aligned}
\tilde{\mathbf{Y}}_{i}^{I} &= \mathrm{Flatten}\big(\mathrm{Stack}(\{\tilde y_{i,g,p}^{I}\}_{g,p})\big) \in \mathbb{R}^{GP\times d_s},\\
\tilde{\mathbf{Y}}_{i}^{T} &= \mathrm{Flatten}\big(\mathrm{Stack}(\{\tilde y_{i,g,p}^{T}\}_{g,p})\big) \in \mathbb{R}^{GP\times d_s}.
\end{aligned}
\end{equation}

In parallel to global token routing at stage $s$, \name{} performs  {\textbf{local routing}} of anatomy-level vision representations $z_{i,a}^{I}$ and text representations $z_{i,a}^{T}$. Let $\mathrm{FFN^L}(\cdot)$ denote the intra-modality expert module processing local anatomy input from both modalities. We form the anatomy stream:
\begin{equation}
\begin{aligned}
\tilde y_{i,a}^{I} = \mathrm{FFN}^{L}_f(z_{i,a}^{I}),
\tilde y_{i,a}^{T} = \mathrm{FFN}^{L}_f(z_{i,a}^{T}) \in \mathbb{R}^{d_s},
\end{aligned}
\end{equation}

where $f \in \{1, \ldots, F\}$, with $F$ the number of anatomy experts, whose outputs are weighted by a learned softmax router shared across modalities. Stacking $\tilde y_{i,a}^{I}$ and $\tilde y_{i,a}^{T}$ over $a$, we obtain $\tilde{\mathbf{Y}}_{i,a}^{I}$ and $\tilde{\mathbf{Y}}_{i,a}^{T} \in \mathbb{R}^{A\times d_s}$ respectively. These resulting matrices provide anatomy context that is concatenated with the global routing output matrices $\tilde{\mathbf{Y}}_{i}^{I}$, $\tilde{\mathbf{Y}}_{i}^{T}$:
\begin{equation}
\mathbf{Q}_i = \mathrm{Concat}\big[\, \tilde{\mathbf{Y}}_{i}^{I};\ \tilde{\mathbf{Y}}_{i}^{T};\ \tilde{\mathbf{Y}}_{i,a}^{I};\ \tilde{\mathbf{Y}}_{i,a}^{T} \big],
\end{equation}
where $\mathbf{Q}_i \in \mathbb{R}^{C_s \times d_s}$ and $C_s$ is the concatenated token length at stage $s$. In practice, the first two stages operate in a vision-only mode, while the last two stages use both vision and text representations. Anatomy masks and text representations are required during pre-training only. At fine-tuning and inference, the model operates on the image alone.

\noindent
\textbf{Second level routing.} The concatenated token matrix $\mathbf{Q}_i$ is processed by a second routing with the same slot-based structure as the first level global routing: tokens are softly assigned to a set of sample- and group-dependent slots, transformed by experts, and then projected back to token space.
Let $\tilde{\mathbf{Q}}_i \in \mathbb{R}^{C_s \times d_s}$ denote the output of this second level routing.
We then split $\tilde{\mathbf{Q}}_i$ back into the vision and text segments, reshape each segment into group--slot form, and reconstruct vision and text slot outputs $q_{i,g,p}^{I}, q_{i,g,p}^{T} \in \mathbb{R}^{d_s}$ back to token space using the first-level assignment weights:
\begin{equation}
\begin{aligned}
\hat{x}_{i,g,k} &= x_{i,g,k} + \sum_{p=1}^{P} \mathbf{A}_{i,g}^{I}[k,p]\, q_{i,g,p}^{I},\\
\hat{t}_{i,g,k} &= t_{i,g,k} + \sum_{p=1}^{P} \mathbf{A}_{i,g}^{T}[k,p]\, q_{i,g,p}^{T}.
\end{aligned}
\end{equation}
This yields fused vision and text tokens ${\hat{x}_{i,g,k},\hat{t}_{i,g,k}} \in \mathbb{R}^{d_s}$ respectively via residual connections.



\subsubsection{Decoder}
After each \name{} stage, the fused vision tokens are reshaped back to the 3D grid to form the multi-scale feature maps $\hat{\mathbf{F}}_i^{(s)}$, which are passed to a UNETR-based~\cite{hatamizadeh2022unetr, plotka2025mamba} decoder via skip connections for voxel-wise segmentation. At each resolution, the decoder upsamples the features and integrates the corresponding encoder-stage output.
\subsection{Training Objectives}
\label{training_obj}

\name{} jointly optimizes segmentation and hierarchical vision-language contrastive alignment. The objectives are designed to ensure that the learned representations encode both global semantics and anatomy-level grounding.

\noindent
\textbf{Global contrastive learning.} 
Whole-image and whole-report representations $(z_i^I, z_i^T)$ from the first \name{} stage ($s=1$) are aligned using a symmetric cosine InfoNCE loss~\cite{oord2018representation}. We define the global contrastive loss $\mathcal{L}_{\mathrm{ITC}}^{G}$ for directions $(M_1, M_2) \in \{(I, T), (T, I)\}$ as:
\begin{equation}
\label{eq:itc-global}
\mathcal{L}_{\mathrm{ITC}}^{G, M_1 \to M_2} = - \frac{1}{B} \sum_{i=1}^B \log \frac{\exp(\langle z_i^{M_1}, z_i^{M_2} \rangle / \tau)}{\sum_{j=1}^B \exp(\langle z_i^{M_1}, z_j^{M_2} \rangle / \tau)},
\end{equation}
where $\langle \cdot, \cdot \rangle$ denotes cosine similarity and $\tau > 0$ is a temperature parameter. We next take their average by 
$\mathcal{L}_{\mathrm{ITC}}^G 
= 1/2\Big(
\mathcal{L}_{\mathrm{ITC}}^{G,I\to T} +
\mathcal{L}_{\mathrm{ITC}}^{G,T\to I}
\Big).$
This symmetric loss encourages the model to pull together matched image-text pairs while pushing apart mismatched pairs in both directions.

\noindent
\textbf{Anatomy-level contrastive learning.} We apply a symmetric contrastive objective at the anatomy level. Let $(z_{i,a}^I, z_{i,a}^T)$ denote extracted representations where $s=1$ for anatomy $a$ in study $i$, and let $A_i$ be the number of anatomies present in $i$. The contrastive loss for both image-to-text ($I \to T$) and text-to-image ($T \to I$) directions is:
\begin{equation}
\label{eq:itc-local}
\begin{split}
\mathcal{L}_{\mathrm{ITC}}^{L, M_1 \to M_2} ={}& \\
-\frac{1}{\sum_{i=1}^B A_i} \sum_{i=1}^B \sum_{a=1}^{A_i} &\log \frac{\exp(\langle z_{i,a}^{M_1}, z_{i,a}^{M_2} \rangle / \tau)}{\sum_{b=1}^{A_i} \exp(\langle z_{i,a}^{M_1}, z_{i,b}^{M_2} \rangle / \tau)},
\end{split}
\end{equation}
where $(M_1, M_2) \in \{(I, T), (T, I)\}$. As in the global case, we average the two directions to obtain $\mathcal{L}^{L}_{\text{ITC}}$, which aligns image features of specific anatomies with their textual snippets. Each $(i, a)$ serves as an anchor, and other anatomy representations $\{(i, b)\}$ within the same sample are treated as negatives.

\noindent
\textbf{Segmentation learning.} 
The segmentation objective is optimized using a sum of Dice ($\mathcal{L}_{\mathrm{Dice}}$) and cross-entropy ($\mathcal{L}_{\mathrm{CE}}$) losses. The Dice loss encourages segmentation accuracy for each anatomical structure, while cross-entropy loss stabilizes voxel-wise classification.

\noindent
\textbf{Adaptive multi-task weighting.}
We use adaptive multi-task weighting based on the Exponential Moving Average (EMA) of each loss magnitude. Let $\mathcal{L}_{\mathrm{seg}}$, $\mathcal{L}_{\mathrm{ITC}}^G$, and $\mathcal{L}_{\mathrm{ITC}}^L$ denote the segmentation, global contrastive, and anatomy-level contrastive losses, respectively. For brevity, let $\ell \in \{\mathrm{seg}, G, L\}$ and denote $\mathcal{L}_\ell^{(t)}$ as the loss value at training step $t$. We maintain an EMA for each loss as:
\begin{equation}
\mu_\ell^{(t)} = (1-\beta)\,\mu_\ell^{(t-1)} + \beta\,\mathcal{L}_\ell^{(t)},
\end{equation}
where $\beta \in (0,1)$ is the smoothing factor and $\mu_\ell^{(0)} = \mathcal{L}_\ell^{(0)}$. Using a small constant $\varepsilon > 0$ for numerical stability, we compute the inverse-magnitude weights
$\tilde w_\ell = 1/(\mu_\ell + \varepsilon)$. These are then normalized by the segmentation weight to obtain the contrastive weights:
\begin{equation}
w_k = \mathrm{clip}\left(\frac{\tilde w_k}{\tilde w_{\mathrm{seg}}}, \alpha_{\min}, \alpha_{\max}\right), \quad k \in \{G, L\}.
\end{equation}
The function $\mathrm{clip}(\cdot, \alpha_{\min}, \alpha_{\max})$ clamps its argument to the interval $[\alpha_{\min}, \alpha_{\max}]$. The hyperparameters $\alpha_{\min}, \alpha_{\max} > 0$ control the allowed range of the contrastive-to-segmentation loss ratios.

\noindent
\textbf{Overall Objective.} 
Finally, the total loss combines segmentation and contrastive terms with the adaptive weights as $\mathcal{L}_{\mathrm{total}} = \mathcal{L}_{\mathrm{seg}} + w_G\, \mathcal{L}_{\mathrm{ITC}}^G + w_L\, \mathcal{L}_{\mathrm{ITC}}^L$. This encourages representations that are anatomically precise and semantically aligned with the corresponding radiology reports, while automatically balancing the contributions of segmentation and global/anatomy-level contrastive objectives.

%% file: sec/6_experiments.tex
\section{Experiments}
\label{sec:experiments}

\begin{table*}[t]
\centering
\scriptsize
\caption{Comprehensive performance comparison across segmentation, classification, and detection tasks. Segmentation performance is reported in mDSC (\%), while PDAC classification and detection are reported via Sensitivity (\%), Specificity (\%), and mAP (\%). VRAM is measured in gigabytes (GB) ($\downarrow$), and speed in million voxels per second (MV/s) ($\uparrow$). The best results are \textbf{bolded}.}
\label{tab:merged_sota_extended}
\resizebox{\textwidth}{!}{%
\begin{tabular}{lccc|cccc|ccc}
\toprule
\textbf{Method} & \textbf{Params} & \textbf{VRAM} & \textbf{Speed} & \multicolumn{4}{c|}{\textbf{Segmentation (mDSC \%)}} & \multicolumn{3}{c}{\textbf{PDAC Cls. \& Det.}} \\
& \textbf{(M)} & \textbf{(GB)} & \textbf{(MV/s)} & \textbf{PANO} & \textbf{AMOS} & \textbf{KiTS} & \textbf{GEPAR3D} & \textbf{Sens. (\%)} & \textbf{Spec. (\%)} & \textbf{mAP (\%)} \\\midrule
Hermes~\cite{gao2024training}    & 44.5  & 4.1 & 12.9 & 77.6 & 85.3 & 65.7 & 84.0 & 75.4 & 79.1 & 62.3 \\
Swin SMT~\cite{plotka2024swin}  & 170.8 & 3.6 & 14.7 & 73.4 & 81.6 & 47.2 & 84.1 & 77.2 & 80.5 & 64.1 \\
VSmTrans~\cite{liu2024vsmtrans}   & 47.6  & 3.1 & 26.2 & 74.7 & 84.8 & 54.6 & 84.6 & 78.8 & 81.3 & 65.8 \\
uC 3DU-Net~\cite{huang2025upping} & \textbf{21.7}  & 3.1 & 50.2 & 75.0 & 82.9 & 60.6 & 83.8 & 73.5 & 77.2 & 60.4 \\
Swin UNETR~\cite{tang2022self} & 72.8  & 4.1 & 27.5 & 72.6 & 83.8 & 48.3 & 82.6 & 80.1 & 82.8 & 67.5 \\
SuPreM~\cite{li2024well}    & 62.2  & 4.1 & 27.5 & 74.6 & 86.3 & 57.0 & 80.6 & 81.6 & 84.4 & 69.2 \\
SegMamba~\cite{xing2024segmamba}   & 66.8  & 3.3 & 25.5 & 72.8 & 84.8 & 56.8 & 86.7 & 83.2 & 85.9 & 71.4 \\
MambaHoME~\cite{plotka2025mamba} & 170.1 & 3.7 & 21.0 & 74.1 & 85.5 & 60.1 & 86.1 & 84.7 & 87.1 & 73.8 \\ \midrule
Merlin~\cite{blankemeier2026merlin} & 256.8 & 3.0 & 96.0 & 77.6   & 85.6  & 64.3   & 86.5   & 85.5   & 87.9   & 76.2   \\
fVLM~\cite{shui2025largescale}  & 110.0 & \textbf{0.6} & \textbf{360.0} & 74.6   & 85.2  & 55.4   & 84.1   & 82.3   & 85.1   & 72.4   \\
BTB3D~\cite{hamamci2025better}     & 101.1 & 1.4 & 60.6 & 77.8   & 86.1   & 55.1   & 84.0   & 84.9   & 87.2   & 74.4   \\
\midrule
 \textbf{\name{}} & 177.9 & 3.7 & 20.8 & \textbf{78.7} & \textbf{87.8} & \textbf{66.7} & \textbf{87.7} & \textbf{87.3} & \textbf{89.5} & \textbf{77.2} \\ \bottomrule
\end{tabular}
}
\end{table*}

\subsection{Experimental setup}

\noindent
\textbf{Datasets.} To \textbf{pre-train} our model, we use the publicly available dataset Merlin~\cite{blankemeier2026merlin,bassi2026merlin} comprising 25,494 CT scans from 18,321 patients and their corresponding radiology reports. For \textbf{fine-tuning,} we use publicly available datasets for segmentation, including PANORAMA~\cite{alves2024panorama}, AMOS-CT~\cite{ji2022amos}, and KiTS 2023~\cite{heller2023kits21}. To assess cross-modality transfer and generalization to anatomy outside the pre-training inventory, we additionally evaluate on GEPAR3D~\cite{szczepanski2025gepar3d}. 
We curate two in-house abdominal CT datasets for (i) classification and detection of PDAC, and (ii) qualitative evaluation of kidney lesions against radiology reports (IRB approval no. 9/2023). The details about data preprocessing, pre-training, fine-tuning and test sets are provided in Appendix \ref{appendix:data_preprocessing} and Appendix \ref{appendix:dataset}. The \textbf{implementation details} and \textbf{evaluation metrics} are provided in Appendices~\ref{appendix:implementation_details} and~\ref{appendix:evaluation_metrics}.

\subsection{Comparison with state-of-the-art}


We compare our proposed model, \name{}, with several state-of-the-art methods, including Hermes~\cite{gao2024training}, Swin SMT~\cite{plotka2024swin}, VSmTrans~\cite{liu2024vsmtrans}, uC 3DU-Net~\cite{huang2025upping}, Swin UNETR~\cite{tang2022self}, SuPreM~\cite{li2024well}, SegMamba~\cite{xing2024segmamba}, and MambaHoME~\cite{plotka2025mamba}, which is our baseline. Additionally, we compare \name{} with pre-trained MedVLMs, including Merlin~\cite{blankemeier2026merlin}, fVLM~\cite{shui2025largescale}, and BTB3D~\cite{hamamci2025better}.


\noindent
\textbf{Downstream tasks.} We evaluate \name{} against state-of-the-art methods on segmentation, classification, and detection. For a fair comparison, all methods are pre-trained on the same dataset and fine-tuned identically on each benchmark; segmentation-based methods are pre-trained on images with the same organ masks, while MedVLMs use both image and text. Table~\ref{tab:merged_sota_extended} shows the quantitative performance of \name{} compared to existing baselines. \name{} consistently achieves the highest mDSC across all four benchmarks, PANORAMA, AMOS, KiTS, and GEPAR3D, demonstrating superior generalizability and robustness across diverse anatomical contexts. We evaluate the classification and detection performance against several state-of-the-art baselines on an independent external test set. \name{} outperforms all methods across all evaluation metrics, with significant gains ($p < 0.05$, Wilcoxon signed-rank test). Detailed results are in Appendix~\ref{appendix:quantitative_results}.

\begin{figure*}[t]
    \centering
    \includegraphics[width=\linewidth]{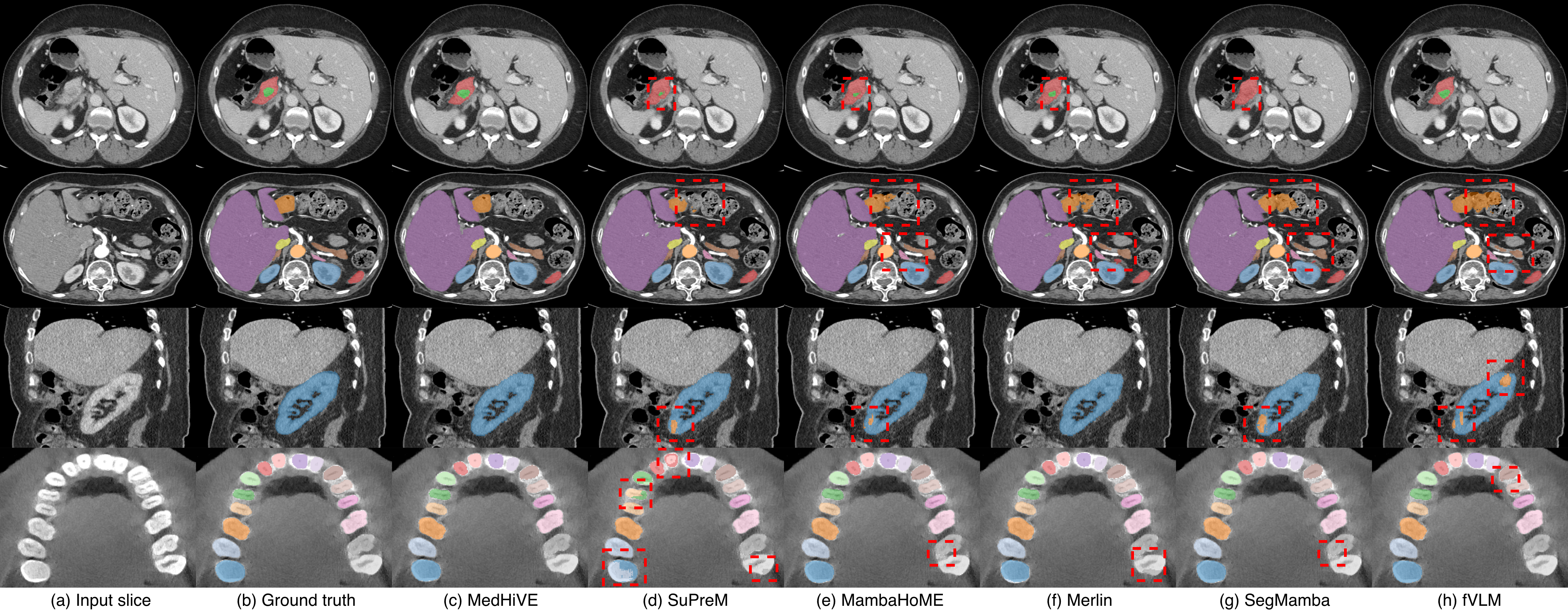}
    \caption{Qualitative segmentation results on four datasets, from top to bottom: PANORAMA, AMOS, KiTS and GEPAR3D. Comparison of \name{} with the top five baselines. Red boxes highlight significant artifacts or missing components in baseline predictions. \name{} shows superior consistency with the ground truth in these challenging regions.}
    \label{fig:main_qualitative_results}
\end{figure*}

\noindent
\textbf{Qualitative results.} We provide a qualitative comparison of \name{} against the top five performing baselines in Figure~\ref{fig:main_qualitative_results}, spanning four diverse datasets. Across all datasets, \name{} demonstrates a superior ability to delineate complex anatomical boundaries and maintain structural integrity in challenging regions. As highlighted by the red boxes, baseline methods frequently exhibit significant artifacts, such as over-segmentation in high-contrast regions or complete failure to capture small-scale lesions. In contrast, \name{} maintains high fidelity to the ground truth, particularly in cases of low-contrast boundaries where typical CNN and Transformer-based architectures produce fragmented or discontinuous predictions. More samples of qualitative results for PANORAMA, AMOS-CT, KiTS, and GEPAR3D are provided in Appendix \ref{appendix:qualitative_results}. We further apply the models fine-tuned on KiTS 2023 to a subset of the in-house cohort with kidney lesions and compare the predicted segmentations against the corresponding radiology reports, as voxel-wise annotations are unavailable for this dataset (see Figure \ref{fig:kits_report}). \name{} produces segmentations that agree with the reported findings, whereas Merlin and fVLM extend beyond lesion boundaries or segment lesions in kidneys the report does not describe, marked by red boxes.

\subsection{Ablation studies}


In this section, we investigate several factors affecting the performance of \name{}: (1) the impact of the key components, including text supervision and local-global contrastive learning; (2) the impact of individual architectural components, including the routing strategy. We pre-train each ablation using the Merlin~\cite{blankemeier2026merlin} dataset and the same training setup. All hyperparameters (e.g., number of experts, groups, and slots) within \layername{} are selected following the ablations reported in MambaHoME~\cite{plotka2025mamba}. The detailed ablation studies are provided in Appendix~\ref{appendix:ablationstudies}.

\begin{table}[t]
\scriptsize
\caption{Ablation studies of the key and  individual architectural components. Segmentation performance (mDSC (\%)) on PANORAMA (PANO), AMOS, KiTS, and GEPAR3D. CL denotes contrastive loss. The best results are \textbf{bolded} and second best results are \underline{underlined}.}
\label{tab:ablation_merged}
\resizebox{\linewidth}{!}{%
\begin{tabular}{l l ccccc}
\toprule
\textbf{\#} & \textbf{Configuration}&
\textbf{PANO} & \textbf{AMOS} & \textbf{KiTS} & \textbf{GEPAR3D} & \textbf{mDSC} \\
\midrule

1 & w/o text \& CL & 74.1 & 85.5 & 60.1 & 86.1 & 76.5 \\
2 & w/o CL &  75.6 & 86.7 & 63.2 & \textbf{88.6} & 78.5\\

3 & w/o SMoE  &  72.8 & 84.8 & 56.8 & 86.7 & 75.3 \\
4 & w/o global CL &  71.2 & 86.1 & 50.0 & 86.2 & 73.4\\
5 & w/o 2nd level routing &  58.2 & 67.1 & 38.7 & 54.1 & 54.5 \\
6 & w/o local routing & 74.2 & 86.8 & 63.7 & 83.9 & 77.2 \\
7 & Text in all stages &  74.6 & 87.5 & 61.1 & 86.4 & 77.4\\
8 & Text in last stage & 74.8 & 86.6 & 65.8  & 81.6  & 77.2\\
9 & Top-\textit{k} MoE &  75.5 & \textbf{87.9} & \textbf{68.3} & 84.2 & 79.0\\
10 & w/o EMA  &  77.1 & 87.6 & 64.6 & 86.9 & \underline{79.1} \\
11 & \textbf{\name{}} &  \textbf{78.7} & \underline{87.8} & \underline{66.7} & \underline{87.7} & \textbf{80.2} \\

\bottomrule
\end{tabular}
}
\end{table}

\noindent
\textbf{Effect of the key components.} Table \ref{tab:ablation_merged} shows that adding text supervision alone already improves performance over the vision-only baseline (\#1), indicating that report-derived semantics provide a complementary signal beyond anatomy masks. Introducing multi-granularity contrastive learning yields further gains (\#2, \#11), supporting our hypothesis that hierarchical alignment is not merely auxiliary supervision, but a mechanism that improves representation quality for downstream dense prediction. Table \ref{tab:amos_ablation_appendix} further shows that gains concentrate where this supervision applies. Averaged over the 12 organs of AMOS covered by our anatomy inventory, mDSC improves by 2.7 from the baseline to full model, compared with 0.7 for the three organs outside it. Even on GEPAR3D, with dental anatomy and CBCT unseen in pre-training, the full model improves over the baseline by 1.6 (\#1, \#11). Similarly, although PDAC and kidney tumors are not pre-training classes, \name{} achieves the highest DSC on both (see Appendix~\ref{appendix:quantitative_results}).

\noindent
\textbf{Effect of the individual architectural choices.} Table \ref{tab:ablation_merged} shows that replacing SMoE with a standard FFN degrades performance, indicating the benefit of dynamic expert routing (\#3). Removing global contrastive alignment also reduces performance, suggesting that global scan–report alignment provides complementary supervision (\#4). Removing second-level routing, where first-level outputs are used directly without fusing, or local routing leads to further drops, highlighting the importance of hierarchical processing (\#5, \#6). The table also shows that enabling text in all stages (\#7) degrades performance. Confining text to the final stage alone (\#8) is also worse than using the last two. This sensitivity to text placement points to a contribution from routing text through \layername{} that goes beyond the contrastive objectives alone. Top-\textit{k} MoE performs comparably on AMOS and slightly better on KiTS, but falls behind on PANORAMA and GEPAR3D (\#9), with soft routing achieving the higher average across the four benchmarks. Using fixed weights, with $w_G$ and $w_L$ both set to 1, does not outperform the adaptive multi-task weighting EMA approach (\#10). Both the hierarchical routing design and multi-level alignment contribute to the performance~(\#11). 


%% file: sec/7_conclusion.tex
\section{Conclusions}
\label{sec:conclusions}


In this work, we present \name{}, a novel hierarchical MedVLM bridging the gap between global clinical semantics and fine-grained 3D anatomical details via a two-level SMoE routing mechanism and multi-granularity contrastive learning. By leveraging organ segmentation, \name{} aligns voxel-level features with corresponding report snippets across anatomical scales, outperforming state-of-the-art global and patch-based models. Ultimately, \name{} provides a scalable foundation for unified 3D medical vision-language understanding, with future work expanding toward longitudinal data.

%% file: sec/acknowledgments.tex
\section*{Acknowledgments}

This research was funded in part by National Science Center, Poland 2025/57/B/ST6/03100. We gratefully acknowledge Polish high-performance computing infrastructure PLGrid (HPC Center: ACK Cyfronet AGH) for providing computer facilities and support within computational grant no. PLG/2025/018792.

%% file: sec/appendix.tex
\clearpage 

\onecolumn

\begin{center}
    {\Large \bfseries Multimodal Hierarchical Mixture-of-Experts\\ for 3D Medical Vision-Language Modeling} \\
    [0.8em]
    
    {\large \bfseries Supplementary Material}
\end{center}

\vspace{1.5em}
\hrule 
\vspace{1.5em}

\appendix

\section{Data Preprocessing}
\label{appendix:data_preprocessing}

\subsection{Segmentation masks}

\begin{figure*}[h]
    \centering
    \includegraphics[width=0.85\textwidth]{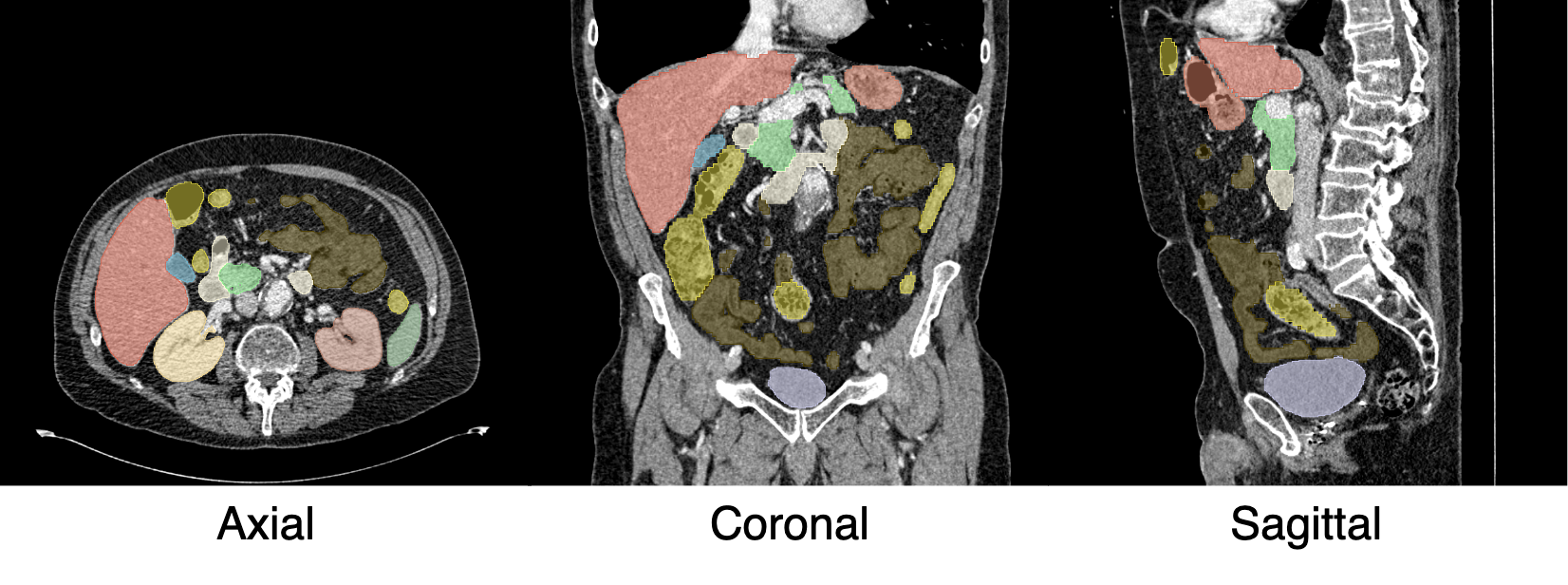}
    \caption{An overview of the 2D projections of anatomical segmentation masks from the Merlin dataset \cite{blankemeier2026merlin}. We consider 14 abdominal anatomical groups to facilitate analysis and model training (see Table~\ref{tab:organs_list}). These projections illustrate the spatial distribution of the anatomies across representative slices.}
    \label{fig:TS_preprocessing}
\end{figure*}

\begin{figure*}[h]
    \centering
    \includegraphics[width=\textwidth, trim={0 0.4cm 0 0}, clip]{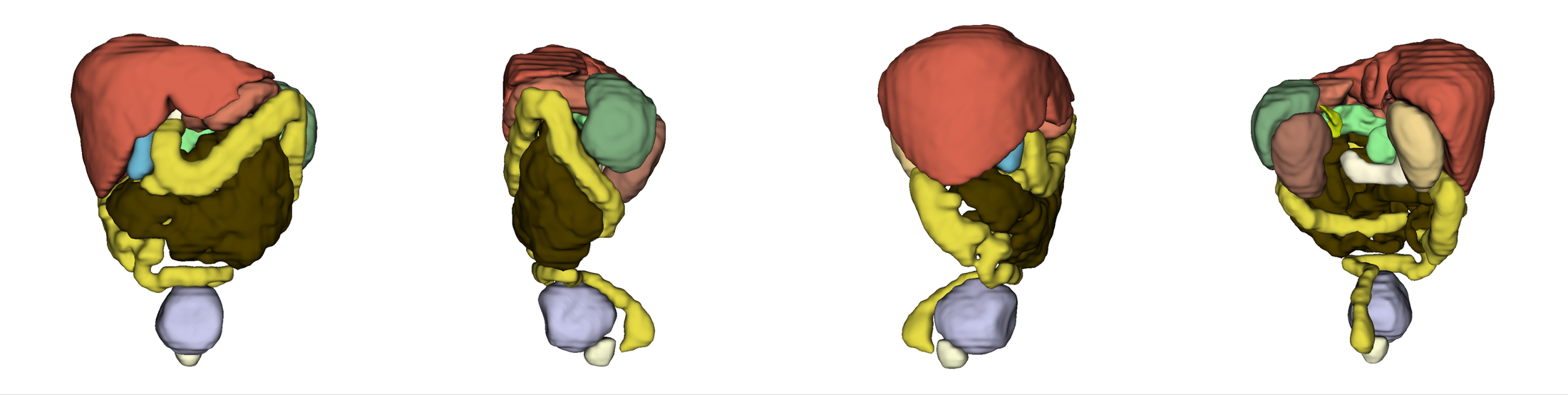}
    \caption{Visualization of the 3D projections of anatomical segmentation masks from the Merlin dataset \cite{blankemeier2026merlin}, shown from the front, left, right, and back views. These projections highlight the global structural consistency and anatomical coherence of the segmented structures. Consistent with the 2D setting, we consider 14 abdominal anatomical groups for clarity and downstream processing (see Table~\ref{tab:organs_list}). The multi-view renderings provide a qualitative assessment of segmentation quality across the full 3D volume.}
    \label{fig:TS_rlfb}
\end{figure*}

We list 14 abdominal anatomical structures within CT scans and match the level of detail in radiology reports, as listed in Table~\ref{tab:organs_list}. The corresponding segmentation masks are sourced from the Merlin dataset \cite{blankemeier2026merlin} and used for pre-training. The masks were obtained through a semi-automatic human-in-the-loop annotation pipeline, in which initial organ masks generated by a pre-trained segmentation model were subsequently refined through iterative quality control and correction procedures. This aligns the fine-grained segmentations with the coarser terminology used in clinical practice, enabling direct comparison with radiologists' descriptions. Furthermore, this process enhances compatibility with standardized reporting frameworks, facilitating downstream analysis and clinical integration. The masks are publicly available as part of Merlin Plus \cite{bassi2026merlin}. Figure~\ref{fig:TS_preprocessing} presents 2D axial, coronal, and sagittal views of these labels, while Figure~\ref{fig:TS_rlfb} provides 3D visualizations.

\subsection{Image-text alignment}
\label{appendix:imagetextalignment}

To enable effective learning from the highly detailed segmentation masks, we directly extract anatomical structures with a more compact and semantically meaningful set of 14 anatomies (see Table~\ref{tab:organs_list}). This consolidation reduces redundancy, improves class balance, and facilitates both visualization and downstream model training by focusing on anatomically coherent regions.

We convert each radiology report into anatomy-specific descriptions that supervise organ-level image and text alignment. We begin by standardizing terminology to a fixed inventory spanning organs, including, for example, the kidney, liver, and pancreas. To make the process robust across institutions and writing styles, each anatomy is associated with a curated set of case-insensitive regular-expression synonyms that cover common lexical variants (e.g., “renal” for kidney), laterality (left/right), lobar names, and frequent spellings. These mappings ensure that heterogeneous phrasing resolves to the same target anatomy.

For each study, we read the \textit{Findings} and \textit{Impression} sections from the report and decide, for each anatomy, whether the report actually discusses it. This “mention” decision combines two signals: a section-conditioned query to Qwen3-30B \cite{yang2025qwen3}, a recent LLM with long-context handling, and a synonym match using compiled regexes. All prompts used to extract reports at the organ-level and global level are provided in Section~\ref{appendix:prompts}.

These anatomy-indexed texts are tokenized to form organ-level representations for local alignment, while the full report produces a global text representation. On the image side, anatomical segmentation masks are used to pool features from the corresponding regions in the 3D CT volume. In combination, anatomy-level image features and anatomy-specific report texts provide local supervision, while the whole-report/whole-image pair supplies global supervision.
\begin{table}[h]
\centering
\caption{List of selected anatomies.}
\begin{tabular}{clcl}
\toprule
No. & Anatomy            & No. & Anatomy            \\
\midrule
1   & Spleen             & 8   & Adrenal gland right \\
2   & Kidney right       & 9   & Adrenal gland left  \\
3   & Kidney left        & 10  & Small bowel         \\
4   & Gallbladder        & 11  & Duodenum            \\
5   & Liver              & 12  & Colon               \\
6   & Stomach            & 13  & Urinary bladder     \\
7   & Pancreas           & 14  & Prostate            \\
\bottomrule
\end{tabular}
\label{tab:organs_list}
\end{table}

\subsection{Text Preprocessing}
\label{appendix:text_preprocess}
\begin{figure*}[t]
\centering

\begin{tcolorbox}[
  width=\textwidth,          
  colframe=reportblue!80,
  colback=reportblue!4,
  coltitle=white,
  fonttitle=\bfseries\footnotesize,
  title=Free-text CT report,
  colbacktitle=reportblue!80,
  fontupper=\small,
]

\textit{FINDINGS:} Lower thorax: Increased attenuation within the lung bases with vascular crowding is likely secondary to atelectasis. 
Liver and biliary tree: Cirrhotic liver morphology with nodularity of the liver surface and relative enlargement of the lateral segment. 
An exophytic enhancing lesion projecting from segment 7 measures 3.5 x 3.1 cm. Another exophytic enhancing lesion along the falciform ligament 
measures 2.1 x 2.0 cm. In segment 2, there is a 2.0 x 1.3 cm enhancing lesion. Gallbladder: Intraluminal gallstones with moderate wall thickening. 
Spleen: Status post splenectomy with multiple splenules and peripheral calcification. Additional enhancing nodules in the right pelvis up to 1.9 cm 
and small enhancing nodules in the omentum up to 1.2 cm. Pancreas: 0.5 cm cystic lesion in the tail; main duct not dilated. Adrenals: Normal. 
Kidneys: No hydronephrosis; left kidney with subcentimeter hypodensities. Small bowel: Mildly dilated fluid-filled loops up to 3.1 cm with collapsed 
distal loops. Colon: Sigmoid diverticulosis; submucosal edema in ascending and transverse colon. Bladder: Moderate wall thickening. Prostate: Normal.
\end{tcolorbox}
\begin{center}
\begin{tikzpicture}[>=Latex]
  \draw[very thick, -{Latex[length=4mm]}, qwenpurple]
    (0,0) -- (0,-1.4);
  \node[right=0.6cm] at (0,-0.7) {%
    \includegraphics[height=6em]{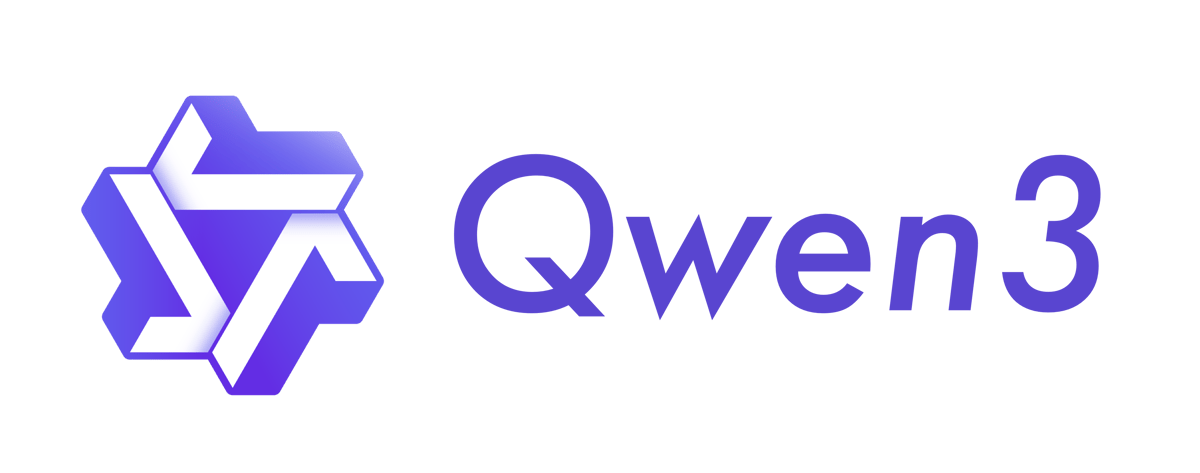}%
    \hspace{0.3em}\scriptsize 
  };
\end{tikzpicture}
\end{center}

\begin{tcolorbox}[
  colframe=jsonteal!80,
  colback=jsonteal!4,
  coltitle=white,
  fonttitle=\bfseries\footnotesize,
  title=Structured output,
  colbacktitle=jsonteal!80,
  fontupper=\small,
  before skip=6pt,
  after skip=3pt
]

\begin{verbatim}
{
  "study_id": "AC4243f52",
  "cleaned_report": "[... The full cleaned report text is here ...]",
  "organs": {
    "Spleen": "Status post splenectomy with multiple splenules. Peripheral 
    calcification around several splenules. Enhancing nodules in the right pelvis 
    up to 1.9 cm. Small enhancing omental nodules up to 1.2 cm.",
    "Kidney right": "Unremarkable.",
    "Kidney left": "Subcentimeter hypodensities in the left kidney.",
    "Gallbladder": "Intraluminal gallstones with moderate gallbladder wall 
    thickening.",
    "Liver": "Cirrhotic morphology with liver surface nodularity and relative 
    enlargement of the lateral segment. Exophytic enhancing lesion in segment 7 
    measuring 3.5 × 3.1 cm. Additional enhancing lesion along the falciform 
    ligament measuring 2.1 × 2.0 cm. Enhancing lesion in segment 2 measuring 
    2.0 × 1.3 cm.",
    "Stomach": "Unremarkable.",
    "Pancreas": "0.5 cm cystic lesion in the pancreatic tail. Main pancreatic duct 
    not dilated.",
    "Adrenal gland right": "Normal.",
    "Adrenal gland left": "Normal.",
    "Small bowel": "Mildly dilated and fluid-filled loops up to 3.1 cm with 
    collapsed distal loops. Wall enhancement not well appreciated.",
    "Duodenum": "Unremarkable.",
    "Colon": "Diverticulosis of the sigmoid colon. Prominent submucosal edema in 
    the ascending and transverse colon.",
    "Urinary bladder": "Moderate bladder wall thickening.",
    "Prostate": "Normal."
  }
}
\end{verbatim}
\end{tcolorbox}
\caption{Text preprocessing pipeline.}
\label{fig:preprocessing}
\end{figure*}

We convert each free-text radiology report into a structured, organ-indexed JSON representation using an LLM \cite{yang2025qwen3}. \Cref{fig:preprocessing} shows a representative example: the raw \textit{Findings} section (top) is parsed into per-organ descriptions (bottom).

\subsection{Prompts}
\label{appendix:prompts}

\noindent
\textbf{Role specification.}
The user message begins by assigning the model the role of a radiology
assistant and constraining its task to per-organ extraction, e.g.
\emph{``You are a highly precise radiology assistant. Extract only current
anatomical findings for each specified organ from the CT report findings
below.''} This line makes clear that the output should consist exclusively of
present-time anatomical observations, organized by organ.

\noindent
\textbf{Critical extraction rules.}
We then list a compact set of high-level extraction constraints that govern the
content of each organ description:
\begin{itemize}
    \item Only \emph{current} findings are allowed; any comparative, temporal,
    or historical statements are to be discarded.
    \item The organ name itself should not be repeated at the beginning of the
    description to avoid redundant phrases like ``Liver: Liver is
    cirrhotic''.
    \item If an organ is not mentioned in the report, the model must output the
    token \texttt{Unremarkable} for that entry.
    \item The text should be short, descriptive, and focused on clinically
    relevant morphology or pathology, with aggressive removal of side comments
    and non-current information.
\end{itemize}

\noindent
\textbf{Filtering rules for non-current information.}
Next, we explicitly enumerate classes of phrases that must be removed
entirely before forming the final snippets:
\begin{itemize}
    \item \emph{Comparative language}, such as ``stable'', ``unchanged'',
    ``no change'', ``increased'', ``interval decrease'', or ``compared to
    prior''.
    \item \emph{Modality references}, for example mentions of PET/CT, MRI,
    ultrasound, X-ray, or other imaging studies and their dates.
    \item \emph{Temporal expressions}, including explicit dates, years, and
    phrases like ``currently'', ``at this time'', or ``now''.
    \item \emph{Bracketed content}, i.e.\ any text in parentheses, square
    brackets, braces or angle brackets, including HU values and measurements
    given inside brackets.
\end{itemize}

\noindent
\textbf{Side-aware rules for bilateral organs.}
Because several structures are bilateral, we additionally condition the model
with side-specific instructions:
\begin{itemize}
    \item When the report refers to a bilateral structure without laterality
    (e.g.\ ``kidneys are normal'', ``both adrenals are unremarkable''), the
    same description is copied to both the right and left entries.
    \item When separate left/right findings are reported (e.g.\ distinct
    cysts in the right and left kidney), the prompt stresses that these must
    remain separated and must not be merged into a single organ description.
    \item For mixed cases, where specific left/right lesions co-occur with
    bilateral changes (e.g.\ ``numerous bilateral hypodense foci''), the model
    is instructed to place the side-specific sentences only in the
    corresponding organ while appending the shared bilateral statement to both.
\end{itemize}

\noindent
\textbf{Cleaning procedure and desired format.}
Finally, the prompt describes a stepwise cleaning procedure (remove bracketed
text, drop comparative and modality phrases, strip dates, discard empty
fragments, default to \texttt{Unremarkable} if nothing remains). The last part of the prompt reiterates that the model must (i) apply the above filtering rules aggressively, (ii) respect the side-aware constraints for kidneys and adrenals, and (iii) return syntactically valid JSON. 

This prompt and conditions provide a JSON block illustrating the output format as shown in  Figure \ref{fig:preprocessing}. The resulting organ-level snippets are then used as $\mathbf{X}_{i,a}^{T}$ for computing anatomy-level text embeddings in the main model.

\subsection{LLM-generated snippet refinement}
\label{appendix:labels_refinement}

Following the automated stages of \cite{bassi2025label}, a dedicated quality control phase is applied to the report-derived snippets. To ensure clinical reliability, all anatomy-aware snippets were manually reviewed by four radiologists with at least 5 years of experience, and refined by a senior radiologist with over 30 years of clinical practice. This verification step focused on preserving anatomical coherence and ensuring that each extracted segment accurately reflects the corresponding content of the original radiology report.

\clearpage

\section{Datasets}
\label{appendix:dataset}

\subsection{Pre-training}

\noindent
\textbf{Merlin.} The Merlin dataset \cite{bassi2026merlin, blankemeier2026merlin} represents one of the largest publicly available repositories for medical imaging research, comprising 25,494 3D computed tomography (CT) scans acquired from a diverse clinical cohort at Stanford Hospital. Unlike traditional segmentation datasets, Merlin is designed for multimodal clinical alignment, pairing high-resolution 3D volumetric data with a massive longitudinal corpus of approximately 1.8 million structured Electronic Health Record (EHR) diagnosis codes and over 6 million tokens of unstructured radiology reports. This integration allows for the development of models that can correlate complex visual features with patient history and clinical outcomes.

The imaging protocols within the dataset reflect real-world clinical workflows, encompassing a variety of contrast phases: the vast majority are portal venous phase scans (97\%), with smaller subsets of delayed (2.4\%), arterial (0.45\%), and non-contrast (0.26\%) studies. Furthermore, the dataset spans over a decade of clinical practice, capturing a wide array of scanner hardware, reconstruction kernels, and slice thicknesses.

\subsection{Fine-tuning}

\noindent
\textbf{PANORAMA.} The PANORAMA dataset \cite{alves2024panorama} comprises 2,238 multi-center contrast-enhanced computed tomography (CECT) scans acquired in a single portal-venous phase. It includes 1,964 newly acquired scans from five European centers, as well as publicly available data from two medical datasets in the United States, namely NIH \cite{roth2015deeporgan} and MSD Pancreas \cite{antonelli2022medical}. The dataset consists of 676 PDAC (Pancreatic Ductal Adenocarcinoma), and 1,562 non-PDAC cases. While the dataset contains six voxel-wise labels, PDAC lesion, veins, arteries, pancreatic parenchyma, pancreatic duct, and common bile duct, we use only the PDAC lesion and merge the pancreatic parenchyma and pancreatic duct classes into a single \textit{pancreas} label. Each CT image consists of 37$\sim$1,572 slices, with a resolution ranging from 512$\times$512 to 1024$\times$1024 pixels. The voxel spatial resolution ranges  ([0.31$\sim$1.03]$\times$[0.31$\sim$1.03]$\times$[0.45$\sim$5.0])mm$^3$ with a mean of 0.75 $\times$ 0.75 $\times$ 2.0 mm$^3$.

\noindent
\textbf{KiTS 2023.} The KiTS dataset \cite{heller2023kits21} represents the third and largest iteration of the Kidney and Kidney Tumor Segmentation challenge, expanding significantly upon the foundations laid in 2019 and 2021. It comprises a total of 599 3D contrast-enhanced CT scans, with 489 cases provided for training and 110 reserved for a hidden test set. While previous versions focused primarily on the late arterial (corticomedullary) phase, KiTS 2023 introduces greater clinical heterogeneity by including scans from the nephrogenic contrast phase, reflecting a broader range of real-world diagnostic protocols. Each scan includes high-fidelity, voxel-level manual annotations for three primary semantic classes: healthy kidney parenchyma, kidney tumors (malignant suspects), and kidney cysts. The annotations were produced through a collaborative effort involving medical students and expert clinicians, ensuring high labeling accuracy even in cases with significant anatomical distortions. The dataset is characterized by extreme variability in target scales; tumor volumes range from small incidentalomas ($0.2 \text{ cm}^3$) to massive renal masses exceeding $2500 \text{ cm}^3$. Geometrically, the scans feature in-plane resolutions between $0.43 \sim 1.04 \text{ mm}$ and slice thicknesses ranging from $0.5$ to $5.0 \text{ mm}$ (with a median of $3.0 \text{ mm}$).

\noindent
\textbf{AMOS-CT.} The AMOS-CT dataset \cite{ji2022amos} consists of a total of 500 CT scans collected from diverse clinical centers, ensuring significant variability in scanner types and patient demographics. However, the ground truth labels are publicly accessible only for the training and validation cohorts, comprising 200 and 100 scans, respectively. Each volume provides high-quality, voxel-level manual annotations for 15 distinct abdominal organs: spleen, right kidney, left kidney, gallbladder, esophagus, liver, stomach, aorta, inferior vena cava, pancreas, right adrenal gland, left adrenal gland, duodenum, bladder, and prostate/uterus. For our experimental setup, we utilize the 200 training scans for model optimization and the 100 validation scans as our test set. The dataset is particularly challenging due to its large intra-class variation; for instance, the number of axial slices per scan ranges from 68 to 353, and the voxel spatial resolution varies significantly from ($[0.45 \sim 1.07] \times [0.45 \sim 1.07] \times [1.25 \sim 5.0]$) $\text{mm}^3$. The mean resolution is $0.70 \times 0.70 \times 4.20$ $\text{mm}^3$. Beyond its scale, AMOS-CT is recognized for including both healthy and pathological cases, requiring models to be robust against anatomical distortions and varied fields of view typical in real-world clinical environments.

\noindent
\textbf{GEPAR3D.} We utilize the GEPAR3D dataset, which is derived from 98 publicly available Cone-Beam Computed Tomography (CBCT) scans originally released by \cite{cui2022fully}. While the initial iteration of this data provided only binary masks (distinguishing tooth structure from background), it presented significant limitations for tasks requiring individual tooth identification or pathological analysis. To address this, \cite{szczepanski2025gepar3d} performed an extensive manual reannotation process. Experienced clinicians labeled the dataset into 32 distinct anatomical classes following the FDI World Dental Federation notation, effectively transforming it into a benchmark for fine-grained multi-class dental segmentation. The dataset is characterized by its high anatomical complexity, featuring diverse dental conditions including restorations and implants that introduce metallic artifacts, varying stages of tooth decay and bone loss, and impacted teeth and irregular dental alignments.

\subsection{External test sets}

\textit{\textbf{Ethical approval for the use of the in-house datasets was granted by the Bioethics Committee of the Faculty of Medicine, Collegium Medicum, University of Warmia and Mazury in Olsztyn (Approval No. 9/2023, March 30, 2023).}}

\subsubsection{Segmentation}

\noindent
\textbf{In-house PDAC CT.} 60 CECT scans of histopathology-confirmed PDAC cases are acquired from 40 medical centers, each containing 68$\sim$875 slices with a voxel spatial resolution ranging ([0.47$\sim$0.98]$\times$[0.47$\sim$0.98]$\times$[0.5$\sim$5.0]) mm$^3$. These scans contain voxel-wise annotations of both the pancreas and the PDAC. The diameter size of the PDAC ranges 1.25$\sim$6.06 cm, with a mean of 3.45$\pm$1.20 cm. All test CECT scans have a resolution of 512 $\times$ 512 pixels. Two junior radiologists annotated the in-house data using 3D Slicer, and the annotations were reviewed by two domain experts with more than 30 years of experience.



\subsubsection{Classification and detection}
\noindent
\textbf{In-house abdominal CT.} We curated a comprehensive and diverse dataset comprising 2,577 CECT abdominal scans obtained from a longitudinal study of 1,740 unique patients, all captured in the portal-venous phase to ensure optimal enhancement of parenchymal organs and venous structures. Within this cohort, the dataset includes 414 cases with histopathology-confirmed Pancreatic Ductal Adenocarcinoma (PDAC), while the remaining cases are treated as a normal cohort for comparative analysis. The dataset’s primary strength lies in its aggregation from 80 distinct medical centers, which introduces a high degree of technical and clinical heterogeneity essential for developing robust and generalizable diagnostic models. By sourcing data from such a wide array of institutions, we account for significant variations in hardware manufacturers and technical acquisition parameters, such as differing tube voltages, current settings, and reconstruction kernels. This multi-center approach also captures a broad range of institutional protocols regarding contrast media injection rates and timing, ensuring the data reflects real-world clinical variability rather than the specific biases of a single facility. We utilize this dataset to validate models for both the classification and detection of PDAC at the patient-level, leveraging the multi-institutional diversity to ensure high diagnostic reliability across varying clinical environments.

\noindent
\textbf{In-house kidney dataset.} We selected a subset of 100 \textit{In-house abdominal CT} scans with kidney tumors and cysts to evaluate our proposed \name{} alongside state-of-the-art methods. Because voxel-wise manually annotated ground-truth segmentations were unavailable for this dataset, our evaluation was performed qualitatively by comparing predictions against the corresponding radiology reports.

\clearpage

\section{Implementation details}
\label{appendix:implementation_details}

All training utilizes the $\mathcal{L}_{total}$ loss function with the AdamW optimizer, an initial learning rate of 1e-4 with a cosine annealing scheduler, a weight decay of 1e-5, and a batch size of 2. Data augmentation is applied on-the-fly, including random rotations, scaling, flipping, and intensity perturbations. Training uses float32 precision.

To improve gradient stability and computational efficiency in SSMs, we use Dynamic Tanh (DyT)~\cite{zhu2025transformers} normalization, defined as:
\begin{equation}
    f_{\text{DyT}}(x) = w \cdot \tanh(\alpha \cdot x) + b,
\end{equation}
where $x \in \mathbb{R}^{B \times d}$ is the input tensor, $w, b \in \mathbb{R}^d$ are learnable per-channel vector parameters, and $\alpha \in \mathbb{R}$ is a shared scalar. DyT applies a point-wise nonlinearity, leveraging the bounded nature of $\tanh$ to stabilize gradients. Unlike Layer Normalization, it avoids costly mean and variance computations, reducing overhead while maintaining $\mathcal{O}(Bd)$ complexity. DyT accelerates training and validation with performance on par with normalization-based alternatives. 

Pre-training is conducted on a workstation equipped with 8 $\times$ NVIDIA H100 GPUs, while fine-tuning and evaluation are performed on a single NVIDIA H100 80GB GPU. Implementation uses Python 3.11, PyTorch 2.4 \cite{paszke2019pytorch}, and MONAI 1.3.0 \cite{cardoso2022monai} within a Distributed Data-Parallel (DDP) training setup.

\subsection{Pre-training}

During pre-training, each CT scan from \textbf{Merlin} is resampled to a spacing of $1.5 \times 1.5\times 1.5\,\text{mm}^3$. All CT scans are clipped to a Hounsfield Unit (HU) window of $[-175, 250]$ and linearly scaled to $[0, 1]$. Pre-training is performed for 100 epochs using patches of size $128 \times 128 \times 128$ voxels as input. During pre-training, we employ on-the-fly augmentations, including scaling, rotation, flipping, adjusting brightness and contrast, and Gaussian smoothing and noise. All baselines are pre-trained on the Merlin dataset using their official implementations and default hyperparameters, for the same number of epochs as MedHiVE. Vision-only methods use images with masks; MedVLMs additionally use the paired radiology reports.

\subsection{Fine-tuning}


\noindent
\textbf{PANORAMA.} Each CT scan is resampled to a spacing of 0.8 $\times$ 0.8 $\times$ 3 mm$^{3}$. All CT scans are clipped to the window of the Hounsfield Unit of \([-175, 250]\) and linearly scaled to \([0, 1]\). During training, we randomly select 1,571 scans (80\%) for training and 393 scans (20\%) for validation, maintaining the distribution ratio of PDAC to non-PDAC cases. During the inference stage, we use a sliding-window algorithm with a patch size of 192 $\times$ 192 $\times$ 48 and an overlap of 0.5 with a Gaussian filter. We train the model for 500 epochs.

\noindent
\textbf{AMOS.} We resample each CT scan to the 1.5 $\times$ 1.5 $\times$ 1.5 mm$^{3}$ isotropic resolution. All CT scans are clipped to the window of the Hounsfield Unit of \([-175, 250]\). Then, we linearly scale to \([0, 1]\). We randomly split the 200 scans from the original training dataset into train and validation subsets with an 80:20 ratio. During the training, we crop random patches of 128 $\times$ 128 $\times$ 128, and a sliding window algorithm with a default overlap of $0.5$ is performed for validation. During training, we employ on-the-fly augmentations, including scaling, rotation, flipping, adjusting brightness and contrast, and Gaussian smoothing and noise. The model is trained for 500 epochs.

\noindent
\textbf{KiTS 2023.} Each CT scan is resampled into the spacing of 0.8 $\times$ 0.8 $\times$ 3 mm$^{3}$ resolution. All CT scans are clipped to the window of the Hounsfield Unit of \([-175, 250]\) and linearly scaled to \([0, 1]\). We randomly split an original training dataset into train and test subsets with 400 and 89 cases, respectively. Since the official KiTS 2023 test set (110 cases) is hidden and not publicly available, all reported results use our own test split drawn from the public training data. Consequently, our Dice scores are not directly comparable to the official challenge leaderboard, which is evaluated on the hidden test set with substantially more training data. During the training, we crop random patches of 128 $\times$ 128 $\times$ 128, and a sliding window algorithm with a default overlap of $0.5$ is performed for validation. During training, we employ on-the-fly augmentations, including scaling, rotation, flipping, adjusting brightness and contrast, and Gaussian smoothing and noise. We fine-tune for 500 epochs.

\noindent
\textbf{GEPAR3D.} We resample each CBCT scan to 0.4 $\times$ 0.4 $\times$ 0.4 mm$^{3}$ isotropic resolution. All CBCT scans are clipped to the window of the Hounsfield Unit of \([0, 5000]\). Then, we linearly scale to \([0, 1]\). We use the official GEPAR3D split, provided by \cite{szczepanski2025gepar3d}. During the training, we crop random patches of 128 $\times$ 128 $\times$ 128, and a sliding window algorithm with a default overlap of $0.5$ is performed for validation. During training, we employ on-the-fly augmentations, including scaling, rotation, flipping, adjusting brightness and contrast, and Gaussian smoothing and noise. Fine-tuning is performed for 1000 epochs.

\clearpage

\section{Evaluation metrics}
\label{appendix:evaluation_metrics}

\noindent
\textbf{Dice Similarity Coefficient (DSC).} The Dice Similarity Coefficient (DSC) serves as our primary measure of regional overlap. Formally, for a multi-class segmentation task with $C$ classes, we represent the prediction and ground truth at each voxel $i$ as one-hot encoded vectors $\mathbf{p}_i, \mathbf{g}_i \in \{0, 1\}^C$. The DSC for class $c$ is defined as the harmonic mean of precision and recall:$$DSC_c = \frac{2 \sum_{i} p_{i,c} g_{i,c}}{\sum_{i} p_{i,c} + \sum_{i} g_{i,c}}.$$The global performance is summarized via the mean Dice Similarity Coefficient (mDSC): $\text{mDSC} = \frac{1}{C} \sum_{c=1}^{C} DSC_c$. Unlike voxel accuracy, mDSC is invariant to the absolute volume of the background, making it particularly effective for medical datasets where the regions of interest (e.g., lesions or small organs) occupy a negligible fraction of the total volume. Mathematically, the DSC can be viewed as an approximation of the $L_1$ distance between the binary masks, normalized by their total cardinality.

\noindent
\textbf{Boundary-based Assessment: 95th Percentile Hausdorff Distance (HD95).} While DSC captures volumetric overlap, it is often insensitive to small, localized boundary deviations which can be clinically significant. To quantify surface alignment, we employ the 95th percentile Hausdorff Distance (HD95). Let $S_c^P$ and $S_c^G$ denote the sets of points on the boundaries of the predicted and ground truth regions for class $c$. The bi-directional Hausdorff distance is typically defined as:
\begin{equation}
d_{H}(S_c^P, S_c^G) = \max\big(
\sup_{p \in S_c^P} \inf_{g \in S_c^G} \| p - g \|_2,
\;\; \sup_{g \in S_c^G} \inf_{p \in S_c^P} \| g - p \|_2 \big).
\end{equation}
However, the standard Hausdorff distance is highly sensitive to outliers—single misclassified voxels far from the ground truth can disproportionately inflate the error. Following standard practice in medical imaging benchmarks, we adopt the 95th percentile of the ordered distances, denoted as $HD95_c$. This robust variant provides a more stable estimate of the ``maximum'' distance between the two surfaces. We report the mean HD95 (mHD95) averaged across all classes to provide a comprehensive view of spatial boundary fidelity.

\noindent
\textbf{Clinical Detection Metrics.} To assess the model's diagnostic utility at the patient level, we evaluate Sensitivity and Specificity. We categorize each patient-level prediction as True Positive ($TP$), False Positive ($FP$), False Negative ($FN$), or True Negative ($TN$) based on the presence of the segmented condition. Sensitivity (True Positive Rate) and Specificity (True Negative Rate) are defined as:$$\text{Sensitivity} = \frac{TP}{TP + FN}, \quad \text{Specificity} = \frac{TN}{TN + FP}.$$These metrics characterize the model’s trade-off between detection coverage and the minimization of false alarms, both of which are critical for clinical integration.

\noindent
\textbf{Patient-level detection.} While segmentation metrics evaluate spatial overlap, mean Average Precision (mAP) assesses the model's ability to localize and rank detections by their confidence scores. For each class $c$, a detection is considered a True Positive ($TP$) if its Intersection over Union (IoU) with a ground truth region exceeds a predefined threshold (typically $IoU \geq 0.5$). We compute the Precision-Recall (PR) curve by varying the detection threshold. The Average Precision (AP) for class $c$ is defined as the area under this curve:$$AP_c = \int_{0}^{1} p(r) \, dr \approx \sum_{n} (r_n - r_{n-1}) p_{interp}(r_n),$$where $p(r)$ is the precision at recall $r$. To provide a global summary across all pathologies, we report the mean Average Precision (mAP), which is the arithmetic mean of $AP$ over all $C$ classes: $\text{mAP} = \frac{1}{C} \sum_{c=1}^{C} AP_c$. This metric is particularly informative for clinical workflows where the ranking of suspicious regions is prioritized to assist radiologist triaging.

\noindent
\textbf{Patient-level classification.} To evaluate the model's diagnostic utility in a clinical setting, we aggregate voxel-wise segmentations into patient-level classifications. A patient is classified as positive if the total predicted volume for a target condition exceeds a predefined threshold $\theta$, and negative otherwise. To characterize the model’s performance in this binary decision space, we report Sensitivity and Specificity.

\clearpage

\section{Quantitative results}
\label{appendix:quantitative_results}

\begin{table*}[h]
\centering
\caption{Quantitative segmentation and detection performance on the PANORAMA test sets (NIH, MSD, and In-house). Evaluation metrics include mDSC (\%) and mHD95 (mm), while PDAC detection efficacy is measured via Sensitivity and Specificity (\%). The best results are highlighted in \textbf{bold}. \name{} demonstrates state-of-the-art performance, particularly in the challenging task of PDAC segmentation, where it significantly outperforms existing Transformer and Mamba-based architectures. Notably, our method achieves a substantial boost in mDSC for the In-house PDAC dataset (65.8\%) and the Overall PDAC average (54.2\%), representing a major improvement over the second-best performing models. Additionally, \name{} exhibits the lowest mHD95 and the highest sensitivity for patient-level detection, underscoring its clinical utility in identifying low-contrast ductal adenocarcinomas across diverse multi-center cohorts.}
\label{tab:panorama_styled_clean}
\scriptsize
\setlength{\tabcolsep}{4pt}
\resizebox{\textwidth}{!}{%
\begin{tabular}{l|ccccc|cc|cc|cc}
\toprule
 & \multicolumn{7}{c|}{\textbf{mDSC (\%) $\uparrow$}} & \multicolumn{2}{c|}{\textbf{mHD95 $\downarrow$}} & \multicolumn{2}{c}{\textbf{Detection $\uparrow$}} \\ \cmidrule(l){2-8} \cmidrule(l){9-10} \cmidrule(l){11-12}
\textbf{Method} & 
\rotatebox{70}{NIH (Pancreas)} & 
\rotatebox{70}{MSD (PDAC)} & 
\rotatebox{70}{MSD (Pancreas)} & 
\rotatebox{70}{In-house (PDAC)} & 
\rotatebox{70}{In-house (Pancreas)} & 
\rotatebox{70}{Overall (PDAC)} & 
\rotatebox{70}{Overall (Pancreas)} & 
\rotatebox{70}{Overall (PDAC)} & 
\rotatebox{70}{Overall (Pancreas)} & 
\rotatebox{70}{Sensitivity} & 
\rotatebox{70}{Specificity} \\ \midrule
Hermes & \textbf{93.6} & 34.8 & \textbf{89.4} & 64.1 & \textbf{82.5} & 52.8 & \textbf{89.1} & 24.5 & 5.1 & 89.0 & 87.9 \\
Swin SMT & 90.2 & 31.0 & 82.3 & 48.0 & 80.4 & 44.5 & 86.1 & 35.8 & 6.8 & 86.6 & 83.5 \\
VSmTrans & 91.0 & 39.5 & 85.4 & 60.2 & 77.7 & 47.2 & 85.4 & 29.4 & 7.8 & 88.0 & 87.2 \\
SegMamba  & 92.1 & 41.1 & 87.9 & 47.2 & 80.2 & 36.2 & 87.6 & 41.1 & 6.3 & 84.8 & 82.4 \\
uC 3DU-Net  & 92.5 & 41.8 & 88.5 & 53.9 & 80.7 & 46.8 & 88.2 & 30.0 & 5.5 & 87.8 & 86.7 \\
Swin UNETR  & 89.2 & 31.3 & 84.0 & 56.1 & 73.7 & 37.0 & 87.1 & 36.8 & 8.3 & 85.4 & 83.4 \\
SuPreM  & 92.0 & 37.8 & 87.7 & 43.3 & 79.1 & 41.2 & 87.3 & 39.8 & 6.2 & 86.4 & 84.1 \\
MambaHoME & 91.3 & 41.3 & 88.3 & 56.8 & 80.5 & 49.2 & 86.7 & 31.3 & 7.1 & 84.6 & 85.3 \\ \midrule
Merlin & 92.4 & 42.5 & 88.3 & 61.8 & 81.0 & 49.4 & 88.0 & 28.2 & 5.9 & 88.3 & 87.0 \\
fVLM & 91.8 & 40.2 & 87.9 & 58.6 & 80.6 & 46.0 & 87.4 & 31.0 & 6.1 & 87.7 & 86.3 \\
BTB3D & 92.7 & 44.2 & 88.8 & 63.2 & 81.2 & 50.6 & 88.5 & 26.9 & 5.7 & 88.7 & 87.1 \\ \midrule
\textbf{\name{}} & 93.1 & \textbf{45.5} & 89.0 & \textbf{65.8} & 81.4 & \textbf{54.2} & 88.9 & \textbf{22.1} & \textbf{4.8} & \textbf{89.5} & \textbf{88.6} \\
\bottomrule
\end{tabular}
}
\end{table*}

\begin{table*}[h]
\centering
\scriptsize
\caption{Quantitative segmentation results on the KiTS dataset. Performance is evaluated across three distinct labels: Kidneys, Tumor, and Cyst. Results are reported in DSC (\%) for individual classes, with mDSC (\%) and mHD95 (mm) representing the overall average performance. \name{} achieves the highest performance in mDSC and mHD95, consistently outperforming both CNN-based and advanced sequence-modeling baselines. Specifically, the model shows a marked improvement in the segmentation of small, irregular structures compared to pure Mamba or Transformer architectures. The achievement of the lowest mHD95 further validates our method's precision in boundary localization, demonstrating its robustness in handling the high anatomical variability and complex pathologies inherent in kidney tumor imaging. The best results are \textbf{bolded}.}
\label{tab:kits_appendix}
\begin{tabular}{l|ccc|cc}
\toprule
\textbf{Method} & \textbf{Kidneys} & \textbf{Tumor} & \textbf{Cyst} & \textbf{mDSC $\uparrow$} & \textbf{mHD95 $\downarrow$} \\ \midrule
Hermes & 88.0 & 53.0 & 47.9 & 65.7 & 49.5 \\
Swin SMT  & 78.4 & 30.0 & 21.8 & 47.2 & 94.2 \\
VSmTrans & 78.6 & 39.0 & 39.4 & 54.6 & 75.3 \\
uC 3DU-Net & 80.0 & 45.8 & \textbf{52.4} & 60.6 & 59.6 \\
Swin UNETR & 78.1 & 31.5 & 24.6 & 48.3 & 92.1 \\
SuPreM & 76.2 & 47.4 & 39.7 & 57.0 & 64.2 \\
SegMamba & 81.4 & 38.7 & 45.2 & 56.8 & 58.4 \\
MambaHoME & 83.5 & 47.0 & 41.6 & 60.1 & 55.4 \\ \midrule
Merlin & 87.6 & 52.4 & 40.4 & 64.3 & 51.2 \\
fVLM & 80.7 & 42.1 & 35.5 & 55.4 & 62.9 \\
BTB3D & 82.6 & 46.7 & 32.9 & 55.1 & 54.2\\\midrule
 \textbf{\name{}} & \textbf{89.6} & \textbf{53.6} & 48.2  & \textbf{66.7}  & \textbf{48.2} \\ \bottomrule
\end{tabular}
\end{table*}

\begin{table*}[h]
\centering
\caption{Quantitative multi-organ segmentation performance on the AMOS-CT dataset. The evaluation encompasses 15 distinct anatomical structures across a wide range of scales and morphologies. mDSC is reported in (\%) and mHD95 in (mm), with the best results per category indicated in \textbf{bold}. \name{} achieves superior overall performance, demonstrating a high degree of robustness across both large, high-contrast organs (e.g., Liver, Spleen) and smaller, more complex structures (e.g., Adrenal Glands, Duodenum). Notably, our method establishes a new state-of-the-art for the Pancreas and Duodenum, which are traditionally challenging due to their low-contrast boundaries and proximity to the gastrointestinal tract. Furthermore, the significant reduction in mHD95 compared to both Transformer-based and pure Mamba-based baselines indicates a superior ability to preserve global topological consistency while maintaining fine-grained local details. \name{} exhibits competitive or superior Dice scores across the majority of classes, underscoring the generalizability of our proposed architecture.}
\label{tab:organ_ablation_full}
\scriptsize
\setlength{\tabcolsep}{3.5pt}
\resizebox{\textwidth}{!}{%
\begin{tabular}{l|ccccccccccccccc|cc}
\toprule
 & \multicolumn{15}{c|}{\textbf{Organ Classes (\%) $\uparrow$}} & \multicolumn{2}{c}{\textbf{Overall}} \\ \cmidrule(l){2-16} \cmidrule(l){17-18}
\textbf{Method} & 
\rotatebox{70}{Spleen} & \rotatebox{70}{R. Kidney} & \rotatebox{70}{L. Kidney} & \rotatebox{70}{Gallbladder} & \rotatebox{70}{Esophagus} & \rotatebox{70}{Liver} & \rotatebox{70}{Stomach} & \rotatebox{70}{Aorta} & \rotatebox{70}{Inferior Vena Cava} & \rotatebox{70}{Pancreas} & \rotatebox{70}{R. Adrenal Gland} & \rotatebox{70}{L. Adrenal Gland} & \rotatebox{70}{Duodenum} & \rotatebox{70}{Bladder} & \rotatebox{70}{Prostate/Uterus} & 
\textbf{mDSC $\uparrow$} & \textbf{mHD95 $\downarrow$} \\ \midrule
Hermes          & 95.0 & 94.1 & 95.0 & 76.7 & 81.9 & 95.9 & 88.2 & 93.2 & 88.7 & 82.0 & 74.0 & 70.7 & 78.4 & 84.7 & 80.3 & 85.3 & 4.9 \\
Swin SMT       & 94.3 & 93.7 & 93.1 & 71.1 & 77.4 & 94.4 & 82.7 & 92.6 & 86.4 & 77.4 & 71.0 & 69.5 & 72.1 & 74.0 & 74.3 & 81.6 & 26.1 \\
VSmTrans  & 95.7 & 94.4 & 93.8 & 76.6 & 80.1 & 95.2 & 86.2 & 92.9 & 87.9 & 79.9 & 74.2 & 72.5 & 76.4 & 85.9 & 80.5 & 84.8 & 19.3 \\
uC 3DU-Net       & 94.1 & 93.2 & 94.5 & 68.2 & 80.7 & 87.7 & 86.0 & 92.8 & 87.2 & 79.4 & 74.2 & 71.9 & 72.6 & 82.7 & 78.1 & 82.9 & 9.6 \\
Swin UNETR       &  92.5 & 90.9 & 93.3 & 73.5 & 80.8 & 94.2 & 84.9 & 93.0 & 86.6 & 82.7 & 74.0 & 72.6 & 75.6 & 82.6 & 78.9 & 83.8 & 25.9 \\
SuPreM      & 95.3 & 94.5 & 95.3 & 78.4 & 82.5 & 96.9 & 89.2 & 93.4 & \textbf{88.8} & 84.9 & 75.3 & 74.7 & 78.4 & 86.4 & 80.0 & 86.3 & 4.8 \\
SegMamba       & 95.4 & \textbf{95.5} & \textbf{95.4} & 72.9 & 80.5 & 93.9 & 87.6 & \textbf{93.5} & \textbf{88.8} & 82.7 & 74.8 & 73.6 & 75.3 & 81.7 & 80.1 & 84.8 & 17.3 \\
MambaHoME       & 95.6 & 93.4 & 92.6 & 75.9 & 81.9 & 95.9 & 88.6 & 93.3 & 87.6 & 83.7 & 74.2 & 75.0 & 76.7 & 86.2 & \textbf{81.1} & 85.5 & 5.0 \\ \midrule
Merlin & 94.9 & 95.3 & 95.1 & 79.5 & 80.0 & 96.7 & \textbf{90.4} & 93.4 & 88.4 & 84.0 & 74.1 & 71.6 & 77.2 & 84.9 & 79.2 & 85.6 & 29.8 \\
fVLM & 95.4 & 94.8 & \textbf{95.4} & 75.2 & 81.4 & 94.7 & 88.0 & 93.1 & 88.0 & 83.1 & 74.6 & 73.9 & 76.5 & 84.2 & 80.4 & 85.2 & 16.4 \\
BTB3D & 96.0 & 95.2 & 93.8 & 80.2 & 81.8 & \textbf{97.1} & 90.0 & 93.2 & 87.2 & 85.0 & 73.6 & 73.4 & 78.8 & 87.3 & 79.2 & 86.1 & 18.3 \\\midrule
 \textbf{\name{}}       & \textbf{96.3} & 94.4 & 95.3 & \textbf{84.2} & \textbf{82.7} & 97.0 & \textbf{90.4} & \textbf{93.5} & 88.6 & \textbf{86.7} & \textbf{77.2} & \textbf{76.8} & \textbf{82.6} & \textbf{89.7} & 80.8 & \textbf{87.8} & \textbf{3.6} \\
\bottomrule
\end{tabular}
}
\end{table*}

\begin{table*}[h]
\centering
\caption{Detailed quantitative evaluation of tooth segmentation on the GEPAR3D test set. Results are categorized by individual teeth following the FDI World Dental Federation notation across all four quadrants. mDSC is reported in (\%) and mHD95 in (mm), with the best results per category indicated in \textbf{bold}. \name{} achieves the highest overall segmentation accuracy and the lowest mHD95 among all compared methods. The performance is notably consistent across different tooth types, effectively bridging the gap between large molars and smaller incisors. In particular, our model demonstrates superior precision in the upper and lower left quadrants, where it maintains high fidelity for complex occlusal surfaces and inter-proximal spaces. The significant reduction in the average Hausdorff distance suggests that \name{} is less prone to the isolated voxel noise and topological errors that often degrade the clinical utility of baseline segmentation outputs.}
\label{tab:dental_segmentation_final}
\small
\setlength{\tabcolsep}{2.5pt}
\resizebox{\textwidth}{!}{%
\begin{tabular}{l|cccccccc|cccccccc|cccccccc|cccccccc|cc}
\toprule
 & \multicolumn{8}{c|}{\textbf{Upper Right (Quadrant 1)}} & \multicolumn{8}{c|}{\textbf{Upper Left (Quadrant 2)}} & \multicolumn{8}{c|}{\textbf{Lower Left (Quadrant 3)}} & \multicolumn{8}{c|}{\textbf{Lower Right (Quadrant 4)}} & \multicolumn{2}{c}{\textbf{Overall}} \\ \cmidrule(l){2-9} \cmidrule(l){10-17} \cmidrule(l){18-25} \cmidrule(l){26-33} \cmidrule(l){34-35}
\textbf{Method} & 
\rotatebox{70}{18} & \rotatebox{70}{17} & \rotatebox{70}{16} & \rotatebox{70}{15} & \rotatebox{70}{14} & \rotatebox{70}{13} & \rotatebox{70}{12} & \rotatebox{70}{11} & 
\rotatebox{70}{21} & \rotatebox{70}{22} & \rotatebox{70}{23} & \rotatebox{70}{24} & \rotatebox{70}{25} & \rotatebox{70}{26} & \rotatebox{70}{27} & \rotatebox{70}{28} & 
\rotatebox{70}{38} & \rotatebox{70}{37} & \rotatebox{70}{36} & \rotatebox{70}{35} & \rotatebox{70}{34} & \rotatebox{70}{33} & \rotatebox{70}{32} & \rotatebox{70}{31} & 
\rotatebox{70}{41} & \rotatebox{70}{42} & \rotatebox{70}{43} & \rotatebox{70}{44} & \rotatebox{70}{45} & \rotatebox{70}{46} & \rotatebox{70}{47} & \rotatebox{70}{48} & 
\textbf{mDSC $\uparrow$} & \textbf{mHD95 $\downarrow$} \\ \midrule
Hermes &  77.8 & 86.6 & \textbf{93.0} & 86.0 & 83.6 & 80.5 & 91.8 & 88.9 & 
85.0 & 89.3 & 86.1 & 80.5 & 89.9 & 91.2 & 90.5 & 85.9 & 
63.5 & 86.8 & 90.3 & 90.1 & 86.8 & 85.8 & 83.0 & 81.4 & 
78.2 & 86.6 & 92.4 & 87.5 & \textbf{93.8} & 85.4 & 88.7 & 53.9 & 
85.2 & 2.1 \\
Swin SMT &  74.4 & 86.4 & 89.8 & 85.0 & 84.6 & 89.6 & 92.4 & 93.1 & 89.2 & 86.0 & 87.5 & 83.0 & 90.1 & 89.7 & 89.6 & 76.4 & 67.2 & 87.1 & 87.3 & 87.2 & 86.4 & 81.7 & 83.4 & 81.9 & 77.8 & 85.9 & 86.7 & 83.9 & 89.5 & 91.7 & 90.5 & 53.2 & 84.8 & 3.4 \\
VSmTrans & 71.4 & 86.3 & 89.1 & \textbf{86.8} & 85.4 & 90.8 & 91.9 & 93.0 & 92.9 & 90.0 & 85.8 & 79.9 & 90.7 & 90.6 & 92.0 & 79.1 & 63.3 & 85.8 & 87.8 & 88.6 & 86.1 & 85.5 & 83.1 & 79.4 & 72.4 & 87.7 & 91.7 & 84.5 & 90.4 & 89.1 & 90.0 & 53.8 & 85.0 & 5.0 \\
uC 3DU-Net & 78.8 & 88.7 & 89.4 & 81.7 & 81.5 & 85.1 & 91.8 & 91.9 & 88.8 & 85.3 & 85.1 & 83.7 & 87.7 & 90.5 & 90.2 & 79.6 & 61.9 & 83.4 & 86.7 & 86.0 & 86.2 & 85.9 & \textbf{86.6} & 82.1 & 78.5 & \textbf{89.4} & 89.6 & 83.2 & 87.6 & 90.1 & 89.5 & 47.5 & 84.3 & 3.3 \\
Swin UNETR & 73.7 & 81.7 & 88.9 & 86.1 & 81.6 & 86.3 & 92.1 & 86.2 & 
77.1 & 87.2 & 84.2 & 83.5 & 87.8 & 90.4 & 90.9 & 75.2 & 
63.8 & 87.1 & 86.7 & 84.9 & 85.4 & 80.6 & 76.2 & 73.5 & 
72.2 & 84.8 & 88.5 & 87.1 & 89.6 & 86.3 & 87.2 & 48.2 & 
82.5 & 3.7 \\
SuPreM & 78.2 & 85.8 & 90.0 & 82.7 & 76.1 & 83.3 & 87.8 & 81.3 & 
83.7 & 84.6 & 84.2 & 80.7 & 88.5 & 89.0 & 87.8 & 79.1 & 
59.9 & 83.1 & 89.2 & 89.0 & 86.4 & 81.0 & 79.5 & 74.9 & 
70.9 & 79.6 & 87.7 & 81.8 & 92.5 & 91.5 & 90.1 & 54.4 & 
82.5 & 2.4 \\
SegMamba & 83.2 & 85.9 & 92.9 & 86.1 & 85.9 & 90.2 & 93.0 & 93.2 & 92.7 & \textbf{91.7} & 88.7 & 80.8 & 88.9 & 94.4 & 92.1 & 85.6 & 74.4 & 88.2 & 87.8 & 90.2 & 87.5 & 85.9 & 83.9 & \textbf{83.5} & \textbf{78.7} & 87.9 & 92.9 & \textbf{87.9} & 90.6 & 92.0 & 90.0 & \textbf{64.1} & 87.1 & 1.6\\ 
MambaHoME & 78.1 & 88.4 & 88.3 & 84.5 & 77.2 & 90.5 & 92.4 & 90.9 & 86.9 & 89.5 & 86.9 & \textbf{84.0} & 90.4 & 90.7 & 91.6 & \textbf{89.2} & 70.0 & 90.1 & 87.2 & 89.9 & 87.3 & 84.2 & 81.6 & 77.9 & 74.6 & 85.6 & 91.9 & 85.6 & 89.7 & 84.5 & 91.1 & 55.8 & 85.4 & 1.7\\ \midrule
Merlin & 77.1 & 87.8 & 90.8 & 85.1 & 80.7 & 90.8 & 92.9 & 93.2 & 87.1 & 84.2 & \textbf{89.2} & 83.1 & 90.5 & 94.8 & 93.1 & 89.0 & 65.9 & \textbf{91.0} & \textbf{91.7} & \textbf{90.5} & \textbf{88.0} & \textbf{86.4} & 83.5 & 81.2 & 75.6 & 86.4 & \textbf{93.1} & \textbf{87.9} & 91.1 & \textbf{93.3} & \textbf{95.0} & 54.2 & 86.5 & 1.7 \\
fVLM & 83.5 & 88.4 & 89.1 & 86.1 & 83.9 & 86.5 & 89.2 & 92.3 & 88.8 & 83.3 & 82.3 & 82.1 & 90.0 & 94.8 & 92.9 & 89.1 & 71.2 & 85.7 & 86.7 & 88.0 & 85.1 & 81.5 & 83.1 & 75.1 & 72.3 & 81.6 & 91.1 & 86.7 & 92.9 & 88.7 & 90.1 & 54.7 & 84.9 & 4.4 \\
BTB3D & \textbf{83.6} & 87.5 & 88.2 & 85.2 & 83.0 & 85.6 & 88.3 & 91.4 & 87.9 & 82.4 & 81.4 & 81.2 & 89.1 & 93.9 & 92.0 & 89.0 & 68.5 & 84.8 & 85.8 & 87.1 & 84.2 & 80.6 & 82.2 & 74.2 & 71.4 & 80.7 & 90.2 & 85.8 & 92.0 & 87.8 & 89.2 & 53.8 & 84.1 & 4.1\\
\midrule
 \textbf{\name{}} & 82.3 & \textbf{89.2} & 89.4 & 86.5 & \textbf{86.7} & \textbf{94.4} & \textbf{93.4} & \textbf{94.1} & \textbf{93.7} & 89.0 & 88.8 & 83.5 & \textbf{91.3} & \textbf{95.0} & \textbf{93.8} & 89.0 & \textbf{83.3} & 90.7 & 90.6 & 89.8 & 87.5 & 85.1 & 85.5 & 79.8 & 74.0 & 84.5 & 88.8 & 87.5 & 93.2 & 92.1 & 91.4 & 56.0 & \textbf{87.6} & \textbf{0.5} \\
\bottomrule
\end{tabular}
}
\end{table*}

\begin{table*}[h]
\centering
\caption{Quantitative segmentation performance on the ToothFairy2 dataset subset. Evaluation follows the FDI World Dental Federation notation across the upper and lower dental arches. mDSC is reported in (\%) and mHD95 in (mm), with the best results highlighted in \textbf{bold}. \name{} demonstrates a clear state-of-the-art advantage, particularly in maintaining structural continuity across the most challenging anatomical sites. Our method significantly outperforms all baselines in the posterior extremities, specifically the third molars (FDI 18, 28, 38, 48), where standard models typically struggle with severe under-segmentation or total loss of connectivity. Furthermore, \name{} attains the second-best mHD95 among all methods. These results validate the robustness of our approach in handling the high morphological variability and lower image quality characteristic of the ToothFairy2 benchmarks, ensuring precise tooth-by-tooth delineation across the entire oral cavity.}
\label{tab:dental_segmentation_toothfairy}
\small
\setlength{\tabcolsep}{2.5pt}
\resizebox{\textwidth}{!}{%
\begin{tabular}{l|cccccccc|cccccccc|cccccccc|cccccccc|cc}
\toprule
 & \multicolumn{8}{c|}{\textbf{Upper Right (Quadrant 1)}} & \multicolumn{8}{c|}{\textbf{Upper Left (Quadrant 2)}} & \multicolumn{8}{c|}{\textbf{Lower Left (Quadrant 3)}} & \multicolumn{8}{c|}{\textbf{Lower Right (Quadrant 4)}} & \multicolumn{2}{c}{\textbf{Overall}} \\ \cmidrule(l){2-9} \cmidrule(l){10-17} \cmidrule(l){18-25} \cmidrule(l){26-33} \cmidrule(l){34-35}
\textbf{Method} & 
\rotatebox{70}{18} & \rotatebox{70}{17} & \rotatebox{70}{16} & \rotatebox{70}{15} & \rotatebox{70}{14} & \rotatebox{70}{13} & \rotatebox{70}{12} & \rotatebox{70}{11} & 
\rotatebox{70}{21} & \rotatebox{70}{22} & \rotatebox{70}{23} & \rotatebox{70}{24} & \rotatebox{70}{25} & \rotatebox{70}{26} & \rotatebox{70}{27} & \rotatebox{70}{28} & 
\rotatebox{70}{38} & \rotatebox{70}{37} & \rotatebox{70}{36} & \rotatebox{70}{35} & \rotatebox{70}{34} & \rotatebox{70}{33} & \rotatebox{70}{32} & \rotatebox{70}{31} & 
\rotatebox{70}{41} & \rotatebox{70}{42} & \rotatebox{70}{43} & \rotatebox{70}{44} & \rotatebox{70}{45} & \rotatebox{70}{46} & \rotatebox{70}{47} & \rotatebox{70}{48} & 
\textbf{mDSC} & \textbf{mHD95} \\ \midrule
Hermes & 39.9 & 80.1 & \textbf{90.6} & 88.6 & 88.2 & 88.2 & 85.9 & 89.2 & 92.3 & 92.2 & 91.3 & 89.9 & 90.3 & 93.0 & 78.4 & 42.8 & 65.0 & 78.8 & 87.0 & 78.5 & 70.3 & 78.6 & 87.6 & 89.1 & 89.3 & 87.4 & 88.7 & 91.3 & 86.9 & 81.8 & 69.7 & 59.5 & 81.7 & 22.1 \\
Swin SMT & 47.5 & 92.5 & 85.5 & 85.6 & 86.6 & \textbf{93.7} & 87.1 & 93.3 & 93.9 & 91.4 & 93.0 & 92.5 & 92.4 & 92.9 & 84.0 & 40.2 & 62.9 & 80.4 & 81.6 & 84.8 & 77.9 & 68.5 & 83.6 & 80.2 & 87.6 & 89.2 & 90.5 & 89.9 & 93.3 & 85.4 & 70.5 & 65.1 & 82.9 & 23.0 \\
VSmTrans & 45.7 & 91.9 & 85.6 & 88.8 & 85.3 & 93.5 & 86.9 & 94.3 & 93.5 & 90.5 & 93.2 & 92.7 & 92.8 & 92.6 & 85.4 & 37.7 & 68.2 & 82.6 & 74.5 & 88.4 & 90.7 & 84.1 & 86.2 & 81.3 & 88.7 & 89.1 & 92.2 & 93.2 & \textbf{93.8} & 81.4 & 72.1 & 65.8 & 83.9 & 21.0 \\
uC 3DU-Net & 47.8 & 90.9 & 84.7 & \textbf{90.0} & 85.8 & 91.4 & 82.5 & 93.2 & 94.1 & 92.5 & 92.6 & 91.5 & 90.8 & 91.8 & 84.4 & 43.5 & 70.1 & 71.1 & 75.5 & 86.0 & 85.4 & 79.9 & 89.1 & 84.1 & 89.9 & 89.6 & 87.9 & 87.0 & 87.4 & 81.6 & 68.9 & 64.5 & 82.8 & 29.9 \\
Swin UNETR & 42.4 & 91.8 & 85.8 & 88.4 & 92.1 & 91.8 & 86.2 & 90.8 & 93.6 & 92.0 & 90.7 & 92.9 & 91.8 & 89.8 & 72.7 & 41.3 & 65.6 & 75.2 & 76.3 & 89.4 & 90.3 & 82.8 & 86.0 & 81.3 & 87.1 & 86.8 & 89.8 & 91.4 & 93.3 & 80.0 & 71.9 & 64.2 & 82.8 & 26.2 \\
SuPreM & 35.9 & 86.2 & 81.5 & 81.0 & 73.2 & 79.4 & 85.0 & 90.5 & 92.9 & 88.9 & 89.8 & 87.3 & 88.2 & 90.7 & 80.6 & 42.3 & 63.8 & 72.6 & 73.5 & 72.3 & 58.7 & 81.0 & 85.3 & 78.8 & 85.0 & 85.4 & 88.5 & 72.5 & 68.1 & 75.6 & 58.9 & 63.9 & 77.0 & 27.4 \\
SegMamba & 55.1 & 93.4 & 85.7 & 89.0 & 87.4 & 93.5 & 86.2 & 93.9 & 94.2 & \textbf{93.2} & 93.1 & 93.1 & \textbf{94.0} & \textbf{94.5} & 86.2 & 44.9 & 68.1 & 84.3 & 78.5 & 88.5 & 89.6 & 85.9 & \textbf{90.0} & 84.8 & \textbf{90.2} & 88.9 & 90.0 & 92.7 & 93.7 & \textbf{85.6} & 71.5 & 63.2 & 85.4 & 14.8 \\
MambaHoME & 47.7 & \textbf{93.6} & 87.4 & 88.9 & 86.8 & 93.5 & \textbf{87.2} & 94.4 & 93.6 & 92.1 & 92.2 & 92.6 & 93.2 & 94.4 & \textbf{87.0} & 37.8 & 59.7 & 75.5 & 75.8 & 88.1 & 87.2 & 83.0 & 88.9 & 83.8 & 89.6 & 88.1 & 91.5 & \textbf{93.7} & 93.4 & 78.3 & 71.2 & 67.4 & 83.8 & 15.9 \\ \midrule
Merlin & 49.7 & 90.7 & 87.2 & 87.6 & 87.0 & \textbf{93.7} & \textbf{87.2} & 93.2 & 92.8 & 92.8 & \textbf{93.5} & \textbf{93.2} & 93.6 & 92.9 & 86.5 & 52.0 & \textbf{75.9} & 87.9 & 79.9 & \textbf{91.7} & 91.0 & \textbf{87.2} & 88.5 & \textbf{90.6} & 90.1 & 89.4 & 92.3 & 92.9 & \textbf{93.8} & 82.0 & 71.7 & 76.6 & 86.5 & \textbf{4.5} \\
fVLM & 47.3 & 88.4 & 83.7 & 87.8 & 89.8 & 91.9 & 85.4 & 93.7 & 93.9 & 91.6 & 91.7 & 91.1 & 92.0 & 92.8 & 82.9 & 51.0 & 66.5 & 81.9 & 91.9 & 88.7 & 87.2 & 82.3 & 83.8 & 78.8 & 85.3 & 85.6 & 85.6 & 87.2 & 90.2 & 74.0 & 64.8 & 68.8 & 83.0 & 28.8 \\
BTB3D & 47.8 & 88.9 & 84.2 & 88.3 & 90.3 & 92.3 & 85.8 & 94.1 & 94.3 & 92.0 & 92.1 & 91.5 & 92.4 & 93.2 & 83.3 & 51.4 & 66.9 & 82.3 & \textbf{92.3} & 89.1 & 87.6 & 82.7 & 84.2 & 79.2 & 85.7 & 86.0 & 86.0 & 87.6 & 90.6 & 74.4 & 65.2 & 69.2 & 83.8 & 28.5 \\ \midrule
 \textbf{\name{}} & \textbf{71.8} & 93.5 & 85.4 & 89.4 & \textbf{93.1} & 93.3 & 86.6 & \textbf{94.6} & \textbf{94.7} & 92.9 & 93.4 & 92.3 & 92.7 & 88.4 & 83.7 & \textbf{57.8} & \textbf{75.9} & \textbf{88.0} & 84.9 & 91.2 & \textbf{91.9} & 86.4 & 89.4 & 90.1 & 90.0 & \textbf{90.2} & \textbf{93.0} & 93.1 & 93.6 & 80.9 & \textbf{76.2} & \textbf{82.0} & \textbf{88.0} & 7.3 \\
\bottomrule
\end{tabular}
}
\end{table*}

\begin{table*}[h]
\centering
\caption{Comparative quantitative analysis of tooth segmentation on the full dataset, including GEPAR3D \cite{szczepanski2025gepar3d} and ToothFairy2 \cite{bolelli2024segmenting,bolelli2025segmenting,lumetti2024enhancing} subsets. Evaluation is conducted for individual teeth using FDI World Dental Federation notation across all four quadrants. mDSC is reported in (\%) and mHD95 in (mm), with the best results \textbf{bolded}. \name{} achieves the highest overall mDSC, particularly in the segmentation of molars and premolars. MedHiVE achieves the second-best mHD95, indicating precise boundary localization. Notably, the performance gap is most pronounced in the posterior regions, where complex root morphologies and proximity to distal anatomical landmarks typically challenge standard segmentation models. These results underscore the effectiveness of our architectural design in maintaining local precision while capturing the global context of the complete dental arch.}
\label{tab:dental_segmentation_combined}
\small
\setlength{\tabcolsep}{2.5pt}
\resizebox{\textwidth}{!}{%
\begin{tabular}{l|cccccccc|cccccccc|cccccccc|cccccccc|cc}
\toprule
 & \multicolumn{8}{c|}{\textbf{Upper Right (Quadrant 1)}} & \multicolumn{8}{c|}{\textbf{Upper Left (Quadrant 2)}} & \multicolumn{8}{c|}{\textbf{Lower Left (Quadrant 3)}} & \multicolumn{8}{c|}{\textbf{Lower Right (Quadrant 4)}} & \multicolumn{2}{c}{\textbf{Overall}} \\ \cmidrule(l){2-9} \cmidrule(l){10-17} \cmidrule(l){18-25} \cmidrule(l){26-33} \cmidrule(l){34-35}
\textbf{Method} & 
\rotatebox{70}{18} & \rotatebox{70}{17} & \rotatebox{70}{16} & \rotatebox{70}{15} & \rotatebox{70}{14} & \rotatebox{70}{13} & \rotatebox{70}{12} & \rotatebox{70}{11} & 
\rotatebox{70}{21} & \rotatebox{70}{22} & \rotatebox{70}{23} & \rotatebox{70}{24} & \rotatebox{70}{25} & \rotatebox{70}{26} & \rotatebox{70}{27} & \rotatebox{70}{28} & 
\rotatebox{70}{38} & \rotatebox{70}{37} & \rotatebox{70}{36} & \rotatebox{70}{35} & \rotatebox{70}{34} & \rotatebox{70}{33} & \rotatebox{70}{32} & \rotatebox{70}{31} & 
\rotatebox{70}{41} & \rotatebox{70}{42} & \rotatebox{70}{43} & \rotatebox{70}{44} & \rotatebox{70}{45} & \rotatebox{70}{46} & \rotatebox{70}{47} & \rotatebox{70}{48} & 
\textbf{mDSC} & \textbf{mHD95} \\ \midrule
Hermes & 64.2 & 84.4 & \textbf{92.1} & 87.0 & 85.1 & 83.2 & 89.7 & 89.0 & 87.7 & 90.3 & 87.9 & 83.8 & 90.1 & 91.8 & 86.3 & 70.5 & 64.1 & 84.0 & \textbf{89.3} & 86.1 & 81.1 & 83.3 & 84.6 & 84.0 & 82.1 & 86.9 & 91.1 & 88.8 & 91.3 & 84.2 & 82.1 & 56.0 & 84.0 & 9.0 \\
Swin SMT & 65.2 & 88.5 & 88.3 & 85.2 & 85.3 & 91.0 & 90.5 & 93.1 & 90.9 & 87.9 & 89.4 & 86.4 & 90.9 & 90.8 & 87.6 & 64.3 & 65.5 & 84.8 & 85.3 & 86.4 & 83.5 & 77.1 & 83.5 & 81.3 & 81.3 & 87.0 & 88.0 & 86.0 & 90.9 & 89.5 & 83.6 & 57.7 & 84.1 & 10.2 \\
VSmTrans & 62.8 & 88.2 & 87.9 & \textbf{87.5} & 85.4 & 91.8 & 90.1 & 93.4 & 93.1 & 90.2 & 88.4 & 84.3 & 91.4 & 91.3 & 89.7 & 64.4 & 65.1 & 84.7 & 83.2 & 88.5 & 87.7 & 85.0 & 84.2 & 80.1 & 78.1 & 88.2 & 91.9 & 87.5 & 91.6 & 86.5 & 83.7 & 58.4 & 84.6 & 10.5 \\
uC 3DU-Net & 67.7 & 89.5 & 87.7 & 84.6 & 83.0 & 87.3 & 88.5 & 92.3 & 90.7 & 87.9 & 87.7 & 86.5 & 88.8 & 91.0 & 88.2 & 66.4 & 64.9 & 79.0 & 82.8 & 86.0 & 85.9 & 83.8 & \textbf{87.5} & 82.8 & 82.6 & \textbf{89.5} & 89.0 & 84.5 & 87.5 & 87.1 & 82.4 & 53.6 & 83.8 & 12.5 \\
Swin UNETR & 62.5 & 85.3 & 87.8 & 86.9 & 85.2 & 88.2 & 90.0 & 87.9 & 83.0 & 88.9 & 86.5 & \textbf{86.8} & 89.2 & 90.2 & 84.6 & 63.6 & 64.4 & 82.9 & 83.3 & 86.5 & 87.1 & 81.4 & 79.6 & 76.2 & 77.4 & 85.5 & 88.9 & 88.6 & 90.8 & 84.1 & 81.9 & 54.0 & 82.6 & 11.5 \\
SuPreM & 62.7 & 85.9 & 87.0 & 82.1 & 75.1 & 81.9 & 86.8 & 84.6 & 87.0 & 86.1 & 86.1 & 83.0 & 88.4 & 89.5 & 85.3 & 66.8 & 61.4 & 79.5 & 83.7 & 83.2 & 76.7 & 81.0 & 81.5 & 76.3 & 75.8 & 81.6 & 88.0 & 78.6 & 83.8 & 85.8 & 79.2 & 58.1 & 80.6 & 11.0 \\
SegMamba & 73.9 & 88.5 & 90.3 & 87.2 & 86.4 & 91.3 & 90.5 & 93.5 & 93.3 & \textbf{92.2} & 90.2 & 85.2 & 90.7 & \textbf{94.4} & 90.0 & 71.1 & 72.0 & 86.9 & 84.6 & 89.6 & 88.2 & 85.9 & 86.0 & 84.0 & 82.8 & 88.2 & 91.8 & 89.6 & 91.6 & \textbf{89.8} & 83.5 & 63.7 & 86.7 & 6.2 \\
MambaHoME & 69.7 & 90.6 & 91.2 & 86.5 & 84.2 & 93.9 & \textbf{91.1} & 93.9 & 91.4 & 92.0 & 89.9 & \textbf{86.8} & 90.9 & \textbf{94.4} & 90.1 & 69.6 & 66.8 & 84.0 & 83.5 & 88.9 & 87.4 & 85.8 & 85.9 & 81.7 & \textbf{84.4} & 89.4 & 92.3 & \textbf{89.8} & \textbf{93.5} & 85.4 & 84.2 & 60.0 & 86.1 & 6.5 \\ \midrule
Merlin & 68.2 & 88.8 & 89.5 & 86.0 & 82.9 & 91.8 & 90.9 & 93.2 & 89.1 & 87.2 & \textbf{90.7} & 86.6 & 91.6 & 94.1 & \textbf{90.8} & 76.7 & 69.6 & \textbf{89.9} & 87.5 & \textbf{90.9} & \textbf{89.0} & \textbf{86.7} & 85.2 & \textbf{84.3} & 80.7 & 87.5 & \textbf{92.9} & 89.7 & 92.0 & 89.3 & \textbf{86.9} & 61.9 & 86.5 & \textbf{2.7} \\
fVLM & 71.2 & 88.2 & 87.0 & 86.5 & 85.7 & 88.1 & 87.6 & 92.6 & 90.4 & 86.0 & 85.3 & 85.1 & 90.4 & 93.8 & 89.2 & 76.7 & 68.1 & 84.1 & 88.1 & 88.0 & 85.6 & 81.5 & 83.1 & 76.2 & 76.6 & 82.8 & 89.0 & 86.8 & 91.8 & 83.5 & 81.1 & 59.5 & 84.1 & 10.1 \\ 
BTB3D & 71.0 & 88.0 & 86.8 & 86.3 & 85.5 & 87.9 & 87.4 & 92.4 & 90.2 & 85.8 & 85.1 & 84.9 & 90.2 & 93.6 & 89.0 & 76.5 & 67.9 & 83.9 & 87.9 & 87.8 & 85.4 & 81.3 & 82.9 & 76.0 & 76.4 & 82.5 & 88.7 & 86.5 & 91.5 & 83.2 & 80.8 & 59.2 & 84.0 & 9.8 \\\midrule
 \textbf{\name{}} & \textbf{79.5} & \textbf{90.7} & 88.0 & \textbf{87.5} & \textbf{88.9} & \textbf{94.0} & 91.0 & \textbf{94.3} & \textbf{94.0} & 90.4 & 90.4 & 86.7 & \textbf{91.8} & 92.7 & 90.3 & \textbf{80.0} & \textbf{80.2} & 89.7 & 88.7 & 90.3 & \textbf{89.0} & 85.5 & 86.9 & 83.3 & 79.6 & 86.5 & 90.2 & 89.5 & 93.3 & 88.1 & 86.1 & \textbf{64.9} & \textbf{87.7} & 3.4 \\
\bottomrule
\end{tabular}
}
\end{table*}

\clearpage

\section{Ablation studies}
\label{appendix:ablationstudies}

\begin{table*}[h]
\centering
\caption{Ablation study on the PANORAMA and in-house datasets demonstrating the impact of multi-modal supervision. We evaluate our proposed method across three diverse test sets: NIH (public), MSD (public), and a multi-institutional In-house dataset. Segmentation quality is measured through mDSC (\%) and mHD95 (mm). Additionally, we report patient-level PDAC detection performance using Sensitivity/Specificity (\%). The results indicate that while text guidance (T) significantly improves tumor localization, the addition of contrastive learning (CL) is crucial for refining boundary transitions, as evidenced by the reduction in mHD95. The best results are \textbf{bolded}.}
\label{tab:panorama_styled_clean_ablation}
\scriptsize
\setlength{\tabcolsep}{4pt}
\resizebox{\textwidth}{!}{%
\begin{tabular}{cccc|ccccc|cc|cc|cc}
\toprule
 & & & & \multicolumn{7}{c|}{\textbf{mDSC (\%) $\uparrow$}} & \multicolumn{2}{c|}{\textbf{mHD95 $\downarrow$}} & \multicolumn{2}{c}{\textbf{Detection $\uparrow$}} \\ \cmidrule(l){5-11} \cmidrule(l){12-13} \cmidrule(l){14-15}
\textbf{\#} & \textbf{B} & \textbf{T} & \textbf{CL} & 
\rotatebox{70}{NIH (Pancreas)} & 
\rotatebox{70}{MSD (PDAC)} & 
\rotatebox{70}{MSD (Pancreas)} & 
\rotatebox{70}{In-house (PDAC)} & 
\rotatebox{70}{In-house (Pancreas)} & 
\rotatebox{70}{Overall (PDAC)} & 
\rotatebox{70}{Overall (Pancreas)} & 
\rotatebox{70}{Overall (PDAC)} & 
\rotatebox{70}{Overall (Pancreas)} & 
\rotatebox{70}{Sensitivity} & 
\rotatebox{70}{Specificity} \\ \midrule

1 & \cmark & \xmark & \xmark & 91.3 & 41.3 & 88.3 & 56.8 & 80.5 & 49.2 & 86.7 & 31.3 & 7.1 & 84.6 & 85.3\\
2 & \cmark & \cmark & \xmark & 92.0 & 44.6 & 88.7 & 61.6 & 81.2 & 51.4 & 88.1 & 28.3 & 5.7 & 87.9 & 87.2\\
 3 & \cmark & \cmark & \cmark & \textbf{93.1} & \textbf{45.5} & \textbf{89.0} & \textbf{65.8} & \textbf{81.4} & \textbf{54.2} & \textbf{88.9} & \textbf{22.1} & \textbf{4.8} & \textbf{89.5} & \textbf{88.6}\\
\bottomrule
\end{tabular}
}
\end{table*}

\begin{table*}[h]
\centering

\caption{Ablation study on the KiTS dataset demonstrating the impact of multi-modal supervision. We evaluate the incremental contribution of text-based prompts (T) and contrastive learning (CL) over the baseline (B) architecture. Segmentation performance is reported via class-specific DSC (\%) for Kidneys, Tumors, and Cysts, alongside aggregate mDSC (\%) and mHD95 (mm) metrics. The results indicate that text guidance (T) primarily benefits cyst segmentation, while the addition of contrastive learning (CL) yields the largest gain for tumors and further reduces mHD95. The best results are \textbf{bolded}.}
\label{tab:kits_appendix_ablation}
\begin{tabular}{lccc|ccc|cc}
\toprule
\# & B & T & CL & Kidneys & Tumor & Cyst & \textbf{mDSC $\uparrow$} & \textbf{mHD95 $\downarrow$} \\ \midrule
1. & \cmark & \xmark & \xmark & 83.5 & 47.0 & 41.6 & 60.1 & 55.4 \\
2. & \cmark & \cmark & \xmark & 86.1 & 48.0 & \textbf{49.1} & 63.2 & 51.0\\
3. & \cmark & \cmark & \cmark & \textbf{89.6} & \textbf{53.6} & 48.2 & \textbf{66.7} & \textbf{48.2} \\ \bottomrule
\end{tabular}
\end{table*}

\begin{table*}[h]
\centering
\caption{Ablation study on the AMOS-CT validation set across 15 organ classes. We evaluate the impact of the Text (T) supervision and Contrastive Loss (CL) components on segmentation accuracy (mDSC) and boundary precision (mHD95). The results demonstrate that incorporating textual representations (Config \#2) consistently improves performance for anatomically complex organs, while the full \name{} framework (Config \#3) achieves the highest overall accuracy and a significant reduction in Hausdorff Distance (mHD95 from 5.0 to 3.6 mm). This suggests that the joint contrastive objective effectively regularizes the latent space, particularly benefiting small-scale structures such as the adrenal glands and duodenum. B, T, and CL indicate baseline, text-conditioned representations, and contrastive loss, respectively. The best results are \textbf{bolded}.}
\label{tab:amos_ablation_full}
\small
\setlength{\tabcolsep}{2.5pt} 
\resizebox{\textwidth}{!}{%
\begin{tabular}{lccc|ccccccccccccccc|cc}
\toprule
 & & & & \multicolumn{15}{c|}{\textbf{Organ Classes (DSC \%)}} & \multicolumn{2}{c}{\textbf{Overall}} \\ \cmidrule(l){5-19} \cmidrule(l){20-21}
\textbf{\#} & \textbf{B} & \textbf{T} & \textbf{CL} & 
\rotatebox{70}{Spleen} & \rotatebox{70}{R. Kidney} & \rotatebox{70}{L. Kidney} & \rotatebox{70}{Gallbladder} & \rotatebox{70}{Esophagus} & \rotatebox{70}{Liver} & \rotatebox{70}{Stomach} & \rotatebox{70}{Aorta} & \rotatebox{70}{IVC} & \rotatebox{70}{Pancreas} & \rotatebox{70}{R. Adr. Gland} & \rotatebox{70}{L. Adr. Gland} & \rotatebox{70}{Duodenum} & \rotatebox{70}{Bladder} & \rotatebox{70}{Pr/Ut} & 
\textbf{mDSC} & \textbf{mHD95} \\ \midrule
1. & \cmark & \xmark & \xmark & 95.6 & 93.4 & 92.6 & 75.9 & 81.9 & 95.9 & 88.6 & 93.3 & 87.6 & 83.7 & 74.2 & 75.0 & 76.7 & 86.2 & \textbf{81.1} & 85.5 & 5.0 \\
2. & \cmark & \cmark & \xmark & 96.1 & \textbf{95.6} & 93.5 & 79.2 & \textbf{83.0} & 96.7 & 89.7 & \textbf{93.8} & \textbf{89.4} & 85.0 & 75.3 & 75.6 & 80.1 & 86.6 & 80.4 & 86.7 & 4.5 \\
 3. & \cmark & \cmark & \cmark & \textbf{96.3} & 94.4 & \textbf{95.3} & \textbf{84.2} & 82.7 & \textbf{97.0} & \textbf{90.4} & 93.5 & 88.6 & \textbf{86.7} & \textbf{77.2} & \textbf{76.8} & \textbf{82.6} & \textbf{89.7} & 80.8 & \textbf{87.8} & \textbf{3.6} \\ \bottomrule
\end{tabular}
}
\label{tab:amos_ablation_appendix}
\end{table*}

\begin{table*}[h]
\centering
\caption{Ablation study on the GEPAR3D test set, excluding ToothFairy2 subset across 32 dental classes (FDI notation). We quantify the influence of Textual Supervision (T) and Contrastive Loss (CL) on fine-grained segmentation. The inclusion of text-conditioned representations (Config \#2) yields substantial gains in semantic recognition across the dental arch. Crucially, the full \name{} framework (Config \#3) achieves the highest boundary fidelity, reaching sub-millimeter precision (mHD95: 0.5 mm). This indicates that the contrastive objective mitigates surface errors effectively in anatomically complex regions, such as the lingual surfaces of incisors (FDI 11, 21) and the occlusal pits of molars (FDI 36, 46), where baselines exhibit higher spatial variance. As this dataset covers dental anatomy absent from our pre-training inventory, the anatomy-level contrastive objective is expected to contribute less here. B, T, and CL denote baseline, text-conditioned representations, and contrastive loss, respectively. The best results are \textbf{bolded}.}
\label{tab:dental_segmentation_filled}
\small
\setlength{\tabcolsep}{2.5pt}
\resizebox{\textwidth}{!}{%
\begin{tabular}{cccc|cccccccc|cccccccc|cccccccc|cccccccc|cc}
\toprule
 & & & & \multicolumn{8}{c|}{\textbf{Upper Right (Quadrant 1)}} & \multicolumn{8}{c|}{\textbf{Upper Left (Quadrant 2)}} & \multicolumn{8}{c|}{\textbf{Lower Left (Quadrant 3)}} & \multicolumn{8}{c|}{\textbf{Lower Right (Quadrant 4)}} & \multicolumn{2}{c}{\textbf{Overall}} \\ \cmidrule(l){5-12} \cmidrule(l){13-20} \cmidrule(l){21-28} \cmidrule(l){29-36} \cmidrule(l){37-38}
\textbf{\#} & \textbf{B} & \textbf{T} & \textbf{CL} & 
\rotatebox{70}{18} & \rotatebox{70}{17} & \rotatebox{70}{16} & \rotatebox{70}{15} & \rotatebox{70}{14} & \rotatebox{70}{13} & \rotatebox{70}{12} & \rotatebox{70}{11} & 
\rotatebox{70}{21} & \rotatebox{70}{22} & \rotatebox{70}{23} & \rotatebox{70}{24} & \rotatebox{70}{25} & \rotatebox{70}{26} & \rotatebox{70}{27} & \rotatebox{70}{28} & 
\rotatebox{70}{38} & \rotatebox{70}{37} & \rotatebox{70}{36} & \rotatebox{70}{35} & \rotatebox{70}{34} & \rotatebox{70}{33} & \rotatebox{70}{32} & \rotatebox{70}{31} & 
\rotatebox{70}{41} & \rotatebox{70}{42} & \rotatebox{70}{43} & \rotatebox{70}{44} & \rotatebox{70}{45} & \rotatebox{70}{46} & \rotatebox{70}{47} & \rotatebox{70}{48} & 
\textbf{mDSC $\uparrow$} & \textbf{mHD95 $\downarrow$} \\ \midrule
1 & \cmark & \xmark & \xmark & 78.1 & 88.4 & 88.3 & 84.5 & 77.2 & 90.5 & 92.4 & 90.9 & 86.9 & 89.5 & 86.9 & 84.0 & 90.4 & 90.7 & 91.6 & 89.2 & 70.0 & 90.1 & 87.2 & 89.9 & 87.3 & 84.2 & 81.6 & 77.9 & 74.6 & 85.6 & 91.9 & 85.6 & 89.7 & 84.5 & 91.1 & 55.8 & 85.4 & 1.7 \\ 
2 & \cmark & \cmark & \xmark & \textbf{85.7} & \textbf{89.3} & \textbf{90.4} & \textbf{87.3} & \textbf{87.0} & 91.1 & \textbf{93.4} & \textbf{94.1} & \textbf{93.9} & \textbf{91.2} & 88.3 & \textbf{84.7} & \textbf{91.5} & 94.8 & 93.5 & \textbf{92.0} & 78.3 & \textbf{93.4} & 88.3 & \textbf{90.6} & \textbf{87.7} & \textbf{87.1} & \textbf{87.5} & \textbf{84.7} & \textbf{81.3} & \textbf{91.2} & \textbf{93.8} & \textbf{91.1} & 90.9 & 91.3 & \textbf{93.3} & \textbf{80.5} & \textbf{89.5} & 0.9 \\
 3 & \cmark & \cmark & \cmark & 82.3 & 89.2 & 89.4 & 86.5 & 86.7 & \textbf{94.4} & \textbf{93.4} & \textbf{94.1} & 93.7 & 89.0 & \textbf{88.8} & 83.5 & 91.3 & \textbf{95.0} & \textbf{93.8} & 89.0 & \textbf{83.3} & 90.7 & \textbf{90.6} & 89.8 & 87.5 & 85.1 & 85.5 & 79.8 & 74.0 & 84.5 & 88.8 & 87.5 & \textbf{93.2} & \textbf{92.1} & 91.4 & 56.0 & 87.6 & \textbf{0.5} \\ 
\bottomrule
\end{tabular}
}
\end{table*}

\begin{table*}[h]
\centering
\caption{Ablation study on the GEPAR3D dataset, including ToothFairy2 subset across 32 individual tooth classes (FDI notation). We evaluate the contribution of Text (T) conditioning and Contrastive Loss (CL) on fine-grained dental segmentation. While the baseline (Config \#1) performs adequately on centralized teeth, the integration of textual representations (Config \#2) provides a significant performance boost in boundary precision, reducing mHD95 from 6.5 to 2.0 mm. Notably, the full \name{} framework (Config \#3) demonstrates superior robustness in segmenting distal molars (e.g., FDI 18, 28, 38), which are typically prone to severe imaging artifacts and morphological variation. This confirms that the contrastive objective effectively regularizes the feature space for anatomically rare or obscured structures. This indicates that the benefit of the anatomy-level contrastive objective can depend on overlap between the pre-training anatomy inventory and the target structures.  B, T, and CL denote baseline, text-conditioned representations, and contrastive loss, respectively. The best results are \textbf{bolded}.}
\label{tab:dental_segmentation_filled_ablation}
\small
\setlength{\tabcolsep}{2.5pt}
\resizebox{\textwidth}{!}{%
\begin{tabular}{cccc|cccccccc|cccccccc|cccccccc|cccccccc|cc}
\toprule
 & & & & \multicolumn{8}{c|}{\textbf{Upper Right (Quadrant 1)}} & \multicolumn{8}{c|}{\textbf{Upper Left (Quadrant 2)}} & \multicolumn{8}{c|}{\textbf{Lower Left (Quadrant 3)}} & \multicolumn{8}{c|}{\textbf{Lower Right (Quadrant 4)}} & \multicolumn{2}{c}{\textbf{Overall}} \\ \cmidrule(l){5-12} \cmidrule(l){13-20} \cmidrule(l){21-28} \cmidrule(l){29-36} \cmidrule(l){37-38}
\textbf{\#} & \textbf{B} & \textbf{T} & \textbf{CL} & 
\rotatebox{70}{18} & \rotatebox{70}{17} & \rotatebox{70}{16} & \rotatebox{70}{15} & \rotatebox{70}{14} & \rotatebox{70}{13} & \rotatebox{70}{12} & \rotatebox{70}{11} & 
\rotatebox{70}{21} & \rotatebox{70}{22} & \rotatebox{70}{23} & \rotatebox{70}{24} & \rotatebox{70}{25} & \rotatebox{70}{26} & \rotatebox{70}{27} & \rotatebox{70}{28} & 
\rotatebox{70}{38} & \rotatebox{70}{37} & \rotatebox{70}{36} & \rotatebox{70}{35} & \rotatebox{70}{34} & \rotatebox{70}{33} & \rotatebox{70}{32} & \rotatebox{70}{31} & 
\rotatebox{70}{41} & \rotatebox{70}{42} & \rotatebox{70}{43} & \rotatebox{70}{44} & \rotatebox{70}{45} & \rotatebox{70}{46} & \rotatebox{70}{47} & \rotatebox{70}{48} & 
\textbf{mDSC $\uparrow$} & \textbf{mHD95 $\downarrow$} \\ \midrule
1 & \cmark & \xmark & \xmark & 69.7 & 90.6 & \textbf{91.2} & 86.5 & 84.2 & 93.9 & 91.1 & 93.9 & 91.4 & \textbf{92.0} & 89.9 & 86.8 & 90.9 & 94.4 & 90.1 & 69.6 & 66.8 & 84.0 & 83.5 & 88.9 & 87.4 & 85.8 & 85.9 & 81.7 & 84.4 & 89.4 & 92.3 & 89.8 & \textbf{93.5} & 85.4 & 84.2 & 60.0 & 86.1 & 6.5 \\ 
2 & \cmark & \cmark & \xmark & 72.3 & \textbf{90.9} & 88.8 & \textbf{87.7} & \textbf{89.1} & 92.0 & \textbf{91.4} & 94.2 & \textbf{94.0} & \textbf{92.0} & 90.1 & \textbf{87.6} & \textbf{92.3} & \textbf{94.7} & \textbf{91.2} & 76.1 & 78.5 & \textbf{90.9} & 85.6 & \textbf{91.1} & \textbf{89.3} & \textbf{86.2} & \textbf{88.7} & \textbf{84.7} & \textbf{84.5} & \textbf{91.2} & \textbf{93.6} & \textbf{92.0} & 92.0 & 87.5 & \textbf{87.1} & \textbf{82.1} & \textbf{88.6} & \textbf{2.0} \\
3 & \cmark & \cmark & \cmark & \textbf{79.5} & 90.7 & 88.0 & 87.5 & 88.9 & \textbf{94.0} & 91.0 & \textbf{94.3} & \textbf{94.0} & 90.4 & \textbf{90.4} & 86.7 & 91.8 & 92.7 & 90.3 & \textbf{80.0} & \textbf{80.2} & 89.7 & \textbf{88.7} & 90.3 & 89.0 & 85.5 & 86.9 & 83.3 & 79.6 & 86.5 & 90.2 & 89.5 & 93.3 & \textbf{88.1} & 86.1 & 64.9 & 87.7 & 3.4 \\
\bottomrule
\end{tabular}
}
\end{table*}

\clearpage

\section{Qualitative results}
\label{appendix:qualitative_results}

To further validate the empirical performance of \name{}, we provide an extensive qualitative comparison across four diverse medical imaging benchmarks: PANORAMA, AMOS-CT, KiTS 2023, and GEPAR3D. While quantitative metrics offer a statistical overview of performance, the following visualizations demonstrate the model's ability to resolve complex anatomical challenges that standard metrics may not fully capture—specifically regarding topological connectivity and boundary fidelity.

\begin{figure*}[h]
    \centering
    \includegraphics[width=\textwidth]{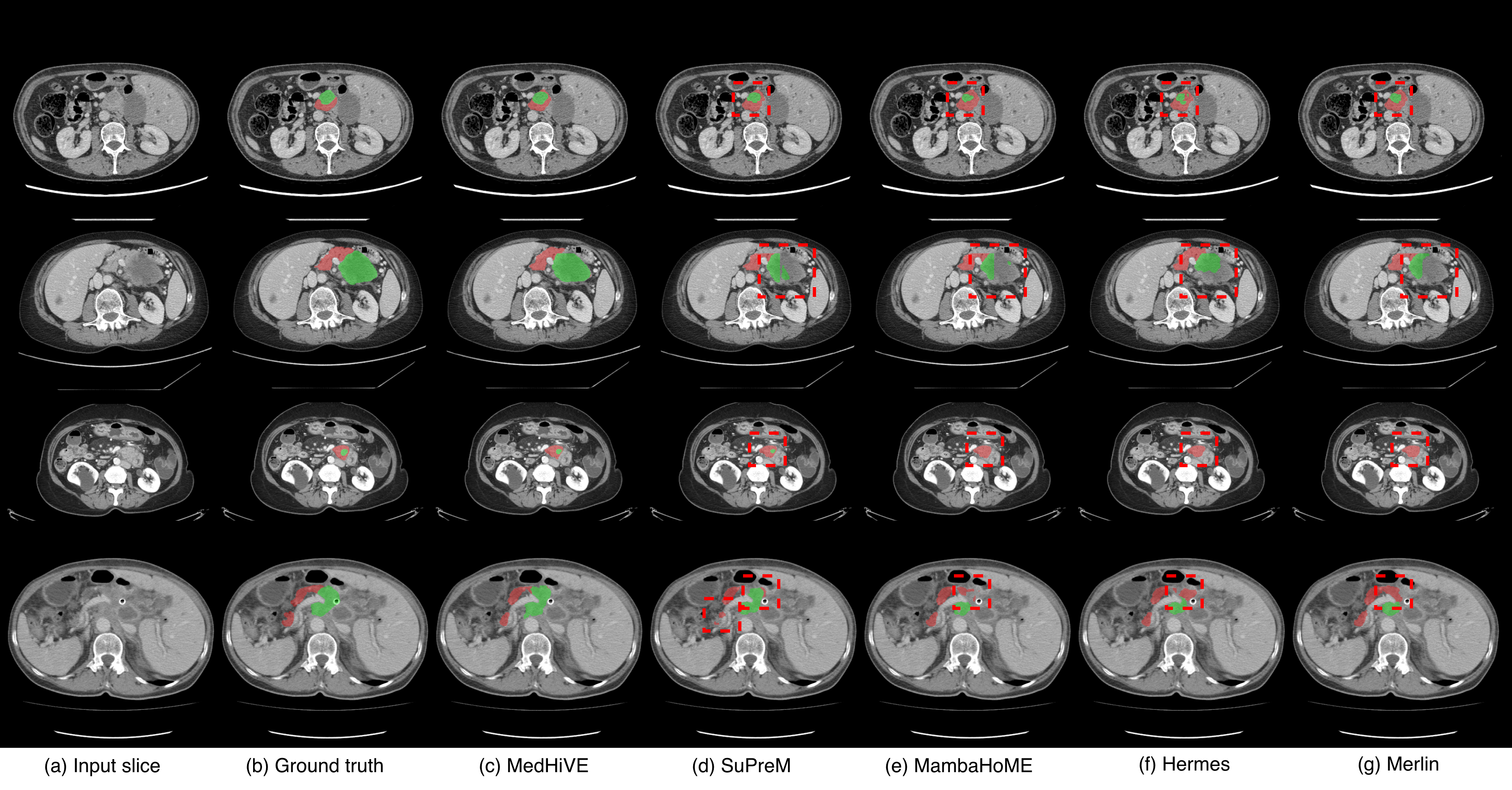}
    \caption{Qualitative segmentation results on the PANORAMA test sets. We compare the proposed \name{} against the ground truth and four state-of-the-art baselines. \name{} demonstrates superior spatial and semantic consistency, accurately capturing long-range dependencies in large-scale structures (e.g., the pancreas parenchyma) while maintaining high-fidelity boundaries for fine-grained objects like PDAC lesions and pancreatic ducts. Unlike competing methods that suffer from topological discontinuities or oversegmentation, our approach consistently preserves structural integrity across varying object scales.}
    \label{fig:appendix_panorama_figure}
\end{figure*}

\begin{figure*}
    \centering
    \includegraphics[width=\textwidth]{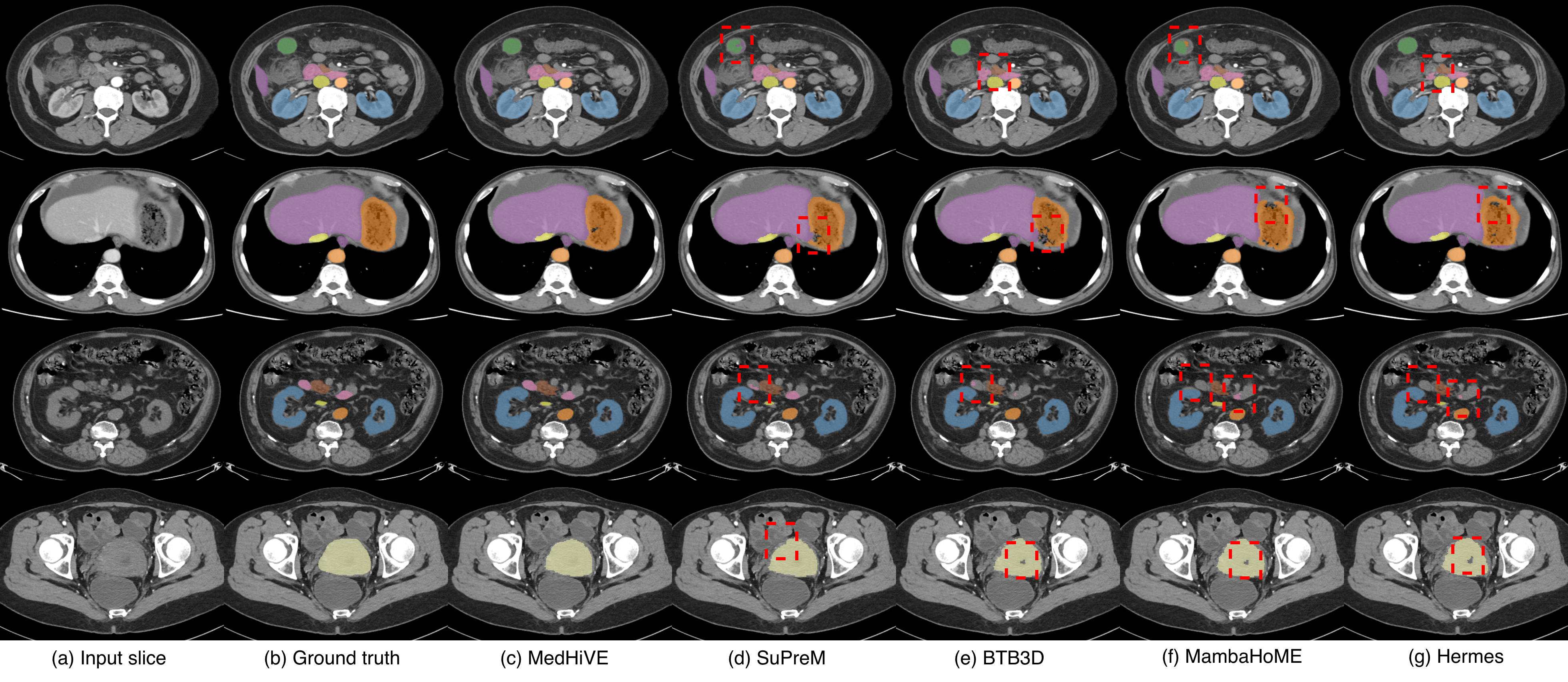}
    \caption{Qualitative multi-organ segmentation performance on the AMOS-CT validation set. We compare the proposed \name{} against ground truth labels and four state-of-the-art baselines across 15 abdominal organs. \name{} demonstrates superior spatial and semantic consistency, leveraging long-range dependency modeling to maintain the global anatomical context of large-scale organs (e.g., liver, stomach) while ensuring high-fidelity boundary localization for small, low-contrast structures (e.g., adrenal glands, gallbladder). While competing methods frequently suffer from inter-class leakage and topological discontinuities in anatomically crowded regions, our approach preserves structural integrity and inter-organ spatial relationships. The highlighted regions (red boxes) illustrate \name{}’s robustness against oversegmentation at organ interfaces, where baseline models often fail to resolve clear semantic boundaries due to similar intensity profiles in non-contrast CT.}
    \label{fig:appendix_amos_figure}
\end{figure*}

\begin{figure*}
    \centering
    \includegraphics[width=\textwidth]{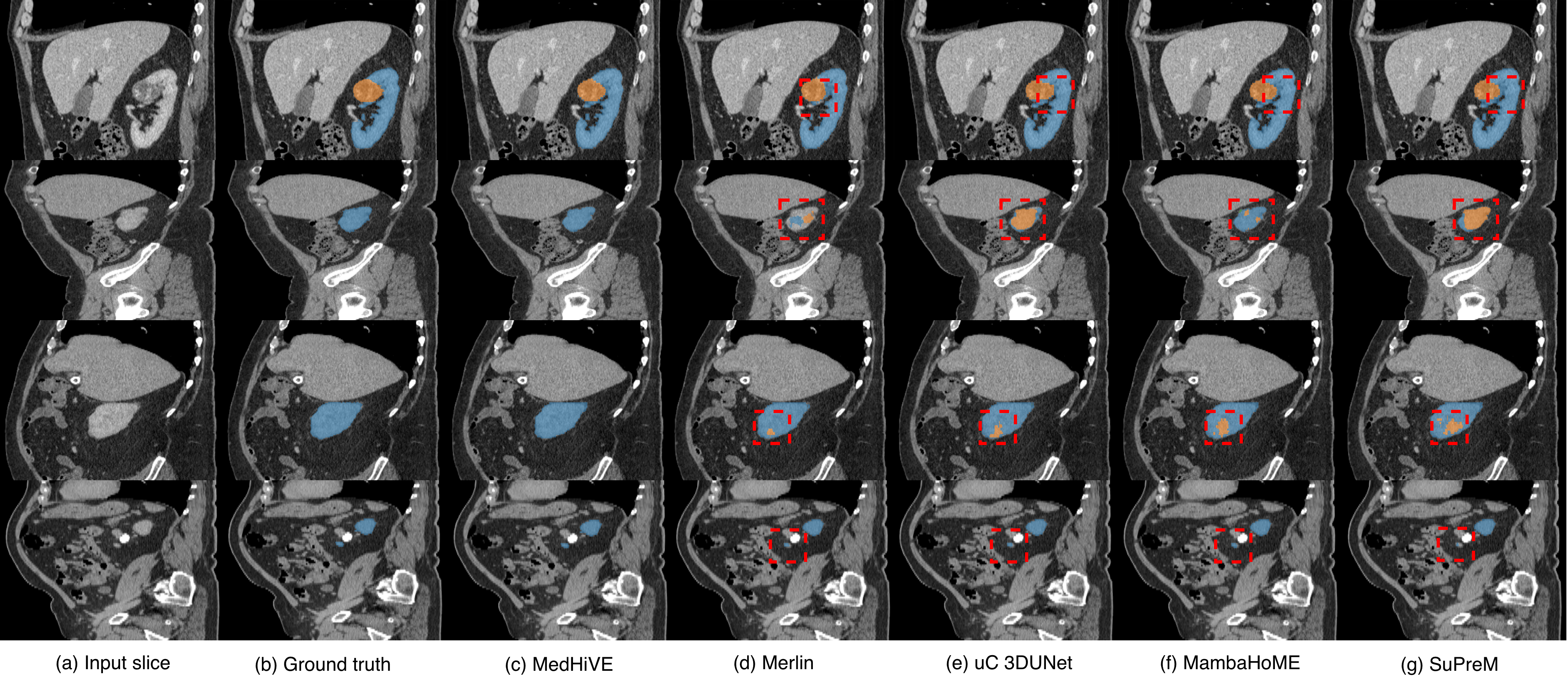}
    \caption{Qualitative multi-class segmentation results on the KiTS 2023 internal test set. We evaluate the proposed \name{} against ground truth annotations and four state-of-the-art baselines for the joint segmentation of the kidney parenchyma, cysts, and tumor masses. \name{} demonstrates superior spatial and semantic consistency by effectively modeling the hierarchical dependencies between nested structures; it captures the global morphology of the kidney while maintaining high-fidelity boundaries for small, low-contrast cysts and heterogeneous tumor regions. Unlike competing methods, which frequently exhibit topological fragmentation or semantic leakage (e.g., misclassifying tumor voxels as healthy tissue), our approach preserves structural integrity across varying object scales. The zoomed-in insets highlight \name{}’s ability to resolve ambiguous boundaries in complex cases of multi-focal cystic disease, where baseline models suffer from oversegmentation and loss of connectivity.}
    \label{fig:appendix_kits_figure}
\end{figure*}

\begin{figure*}
    \centering
    \includegraphics[width=\textwidth]{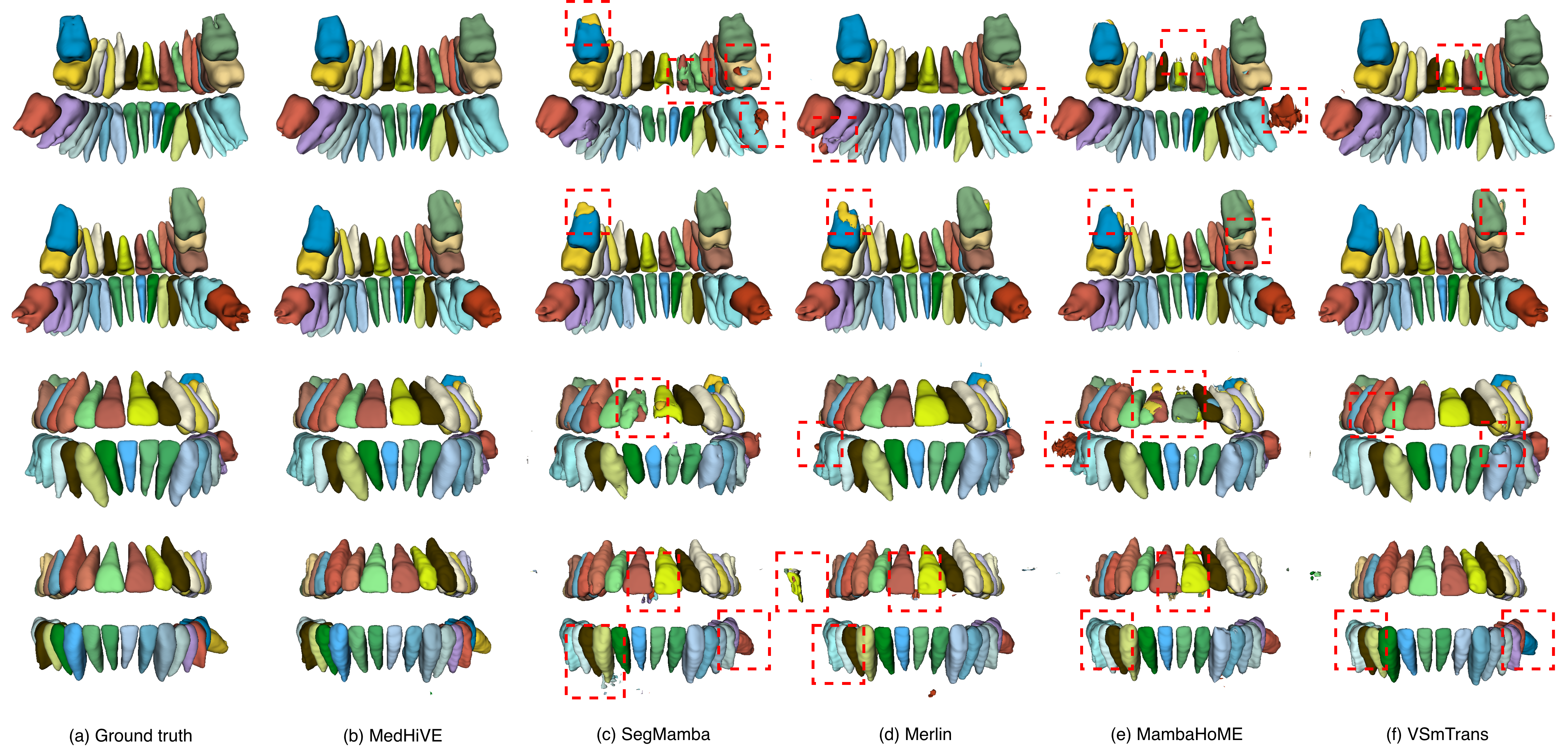}
    \caption{Qualitative evaluation of tooth segmentation on the GEPAR3D test set. We benchmark \name{} against ground truth labels and four state-of-the-art baselines. \name{} exhibits superior spatial and semantic consistency, leveraging its ability to capture long-range dependencies for large-scale dental arches while preserving high-fidelity boundaries in smaller, individual tooth structures. While competing methods frequently suffer from topological discontinuities (e.g., fragmented roots) or oversegmentation in high-density regions, our approach maintains global structural integrity and local connectivity. The zoomed-in regions (red boxes) highlight \name{}'s robustness in resolving ambiguous inter-proximal boundaries where baseline models typically fail to maintain clear separation.}
    \label{fig:appendix_gepar3d_figure}
\end{figure*}

The qualitative results are organized as follows:
\begin{enumerate}
    \item Pancreatic Structures (Figure~\ref{fig:appendix_panorama_figure}): In the PANORAMA dataset, we highlight the model’s ability to maintain the structural integrity of the pancreas parenchyma while simultaneously isolating fine-grained PDAC lesions and ducts,
    \item Abdominal Multi-Organ Anatomy (Figure~\ref{fig:appendix_amos_figure}): We visualize the robustness of our approach in crowded anatomical regions, demonstrating a significant reduction in inter-class leakage between adjacent organs like the liver and stomach,
    \item Nested Kidney Morphologies (Figure~\ref{fig:appendix_kits_figure}): Results from the KiTS 2023 set showcase how \name{} handles hierarchical dependencies, accurately separating kidney parenchyma from cysts and heterogeneous tumor masses without the fragmentation common in baseline methods,
    \item Dental Arch (Figure~\ref{fig:appendix_gepar3d_figure}): Finally, we evaluate the model on the GEPAR3D dataset, emphasizing its precision in resolving ambiguous inter-proximal boundaries and preserving the connectivity of delicate root structures.
\end{enumerate}

Collectively, these visualizations confirm that \name{} consistently outperforms state-of-the-art baselines by capturing long-range dependencies and maintaining high-fidelity semantic boundaries across varying object scales and imaging modalities.

\begin{figure*}
    \centering
    \includegraphics[width=\textwidth]{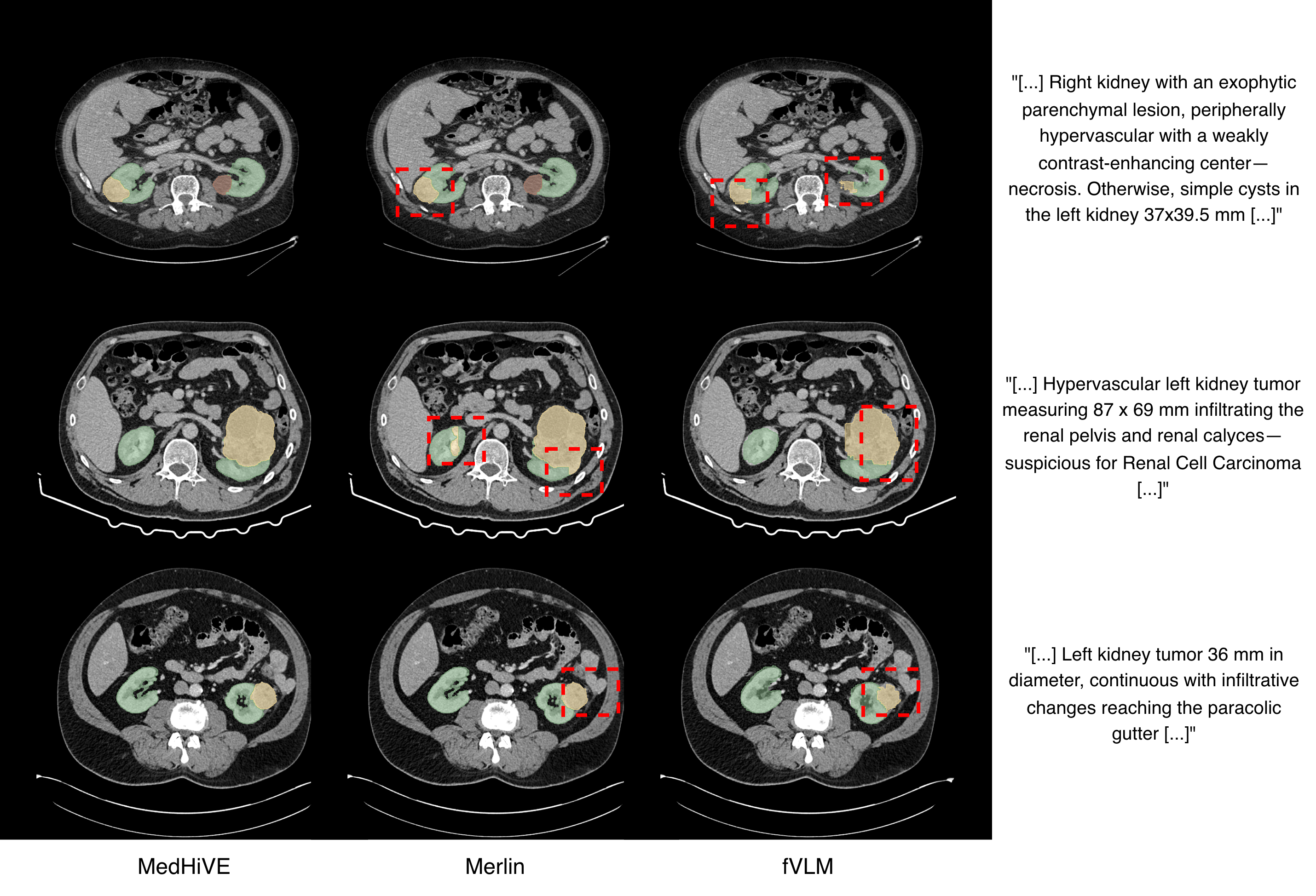}
    \caption{Qualitative comparison of visual grounding performance on external in-house abdominal CT scans with kidney lesions. Each row displays a representative abdominal CT slice paired with its corresponding part of radiology report describing kidney tumors or cysts; in the absence of voxel-wise annotations, the report serves as the reference. Grounding predictions are shown for MedHiVE (left), Merlin (center), and fVLM (right). All models are fine-tuned on KiTS 2023. MedHiVE segments lesions consistent with the radiology reports, whereas Merlin and fVLM overshoot the lesion extent or place predictions in a kidney the report leaves unremarkable (red boxes).}
    \label{fig:kits_report}
\end{figure*}

\clearpage

\section{Clinical impact}

Hierarchical medical vision--language models such as \name{} are best
viewed as \emph{assistive} technology: they learn volumetric features
that cohere with how radiologists describe anatomy in free text, which
can reduce repetitive manual work, speed up research cohort
construction, and improve access to structured imaging cues in
resource-constrained settings. In practice, the same class of model
could support draft contouring for treatment planning, quality assurance
on routine cross-sectional studies, retrieval of similar prior cases,
and teaching tools that link scrollable volumes to report language.
Because \name{} separates local token routing from global
vision--language fusion, downstream systems can in principle highlight
\emph{which} image regions and \emph{which} phrases contribute to a
prediction---a property that matters for auditability and for training
junior readers, provided uncertainty is communicated honestly.

The public benchmarks we evaluate are \emph{examples} of these use
cases, not an exhaustive clinical product specification. Work on
pancreatic ductal adenocarcinoma (PDAC) in portal-venous abdominal CT
illustrates how better multi-structure segmentation could ease contouring
for chemoradiation or surgical planning, while leaving resectability,
margin assessment, and staging decisions to the treating team.
Conversely, fine-grained \emph{three-dimensional} tooth segmentation on
cone-beam CT (as in GEPAR3D) illustrates the same principle outside
abdominal radiology: sharper tooth-level masks can assist orthodontic,
implant, and maxillofacial planning software, but metal artifact,
missing teeth, and atypical anatomy still require interactive human
review. Across domains, any deployment must be preceded by local
validation, clear human-in-the-loop governance, and---where a product is
intended for care---the appropriate regulatory pathway. \name{} has
not been evaluated in a prospective reader study and is not a
certified medical device; it should not be used to make treatment
decisions without expert oversight.

\clearpage

\section{Limitations and future directions}
\label{sec:limitations}

While \name{} demonstrates significant improvements in 3D multi-modal understanding, a few {\textbf{limitations}} remain that offer avenues for future research.

\noindent
\textbf{Fixed anatomical inventory.} The current framework operates on a predefined set of 14 anatomical structures, which covers the most clinically relevant regions in the evaluated datasets. While this design enables stable and interpretable anatomy-level alignment, it may limit coverage of rare or fine-grained structures. Importantly, our results indicate that the learned representations still generalize beyond these regions through global context. Extending the inventory or incorporating more flexible anatomical representations is a direction for future work.

\noindent
\textbf{Data biases.} 
The model is trained on large-scale clinical data, where the frequency of anatomical descriptions follows real-world reporting patterns. As a result, commonly discussed organs are more densely represented than less frequently mentioned regions. While this reflects realistic clinical distributions and does not prevent strong performance across benchmarks, it may influence sensitivity to rare findings. Future work could explore reweighting strategies or targeted sampling to further improve performance on underrepresented anatomies. The development of \name{} opens several {\textbf{promising directions}} for extending hierarchical vision-language modeling in the medical domain.

\noindent
\textbf{Open-vocabulary anatomical grounding.} A primary goal for future iterations is to move beyond a fixed anatomical inventory. By integrating open-vocabulary object detection techniques with our hierarchical routing, the model could potentially align visual tokens with arbitrary clinical concepts mentioned in reports, such as specific lesions or rare vascular anomalies, without requiring predefined masks. This would transition the system from an ``organ-centric'' model to a ``pathology-centric'' one.

\noindent
\textbf{Longitudinal integration of imaging and clinical context.} Clinical diagnosis often relies on the evolution of findings over time. Extending the Mamba-based backbone to handle 4D data (3D volumes over multiple time points) would allow the model to capture disease progression. This would require the \name{} to route information not only across spatial scales but also across temporal sequences, while incorporating tokens derived from longitudinal clinical context beyond radiology reports (e.g., pathology results, operative notes, laboratory trends, or prior diagnoses) to align imaging changes with the evolving clinical narrative.

\noindent
\textbf{Multi-modality validation.} We aim to evaluate the generalizability of the \name{} framework across different imaging modalities, such as MRI and PET-CT, and across diverse patient populations from multiple clinical sites. This would involve studying the domain-shift robustness of the hierarchical routers when encountering different scanner protocols and reporting styles. 

\clearpage

\section{Impact statement}
\label{appendix:impactstatement}

\noindent
\textbf{Contribution.}
This work contributes a hierarchical medical vision--language framework
that couples two-stage Soft Mixture-of-Experts routing with
multi-granularity contrastive alignment between 3D CT and radiology
text. The scientific advance is a unified way to inject anatomical
structure and report structure into the same encoder, improving
downstream segmentation, classification, and detection on public
benchmarks relative to global or patch-only MedVLMs.

\noindent
\textbf{Positive societal impacts.}
Better alignment between imaging and clinical language can reduce
repetitive manual annotation, accelerate research cohort construction,
and improve access to structured imaging cues in settings with
limited specialist time. Open-science release of code, prompts, and
evaluation protocols (where licensing permits) can lower the barrier
for reproducible medical AI.

\noindent
\textbf{Negative societal impacts and mitigations.}
Like all data-driven medical models, \name{} can inherit and amplify
\emph{documentation bias} (who is scanned, which findings are written
down, and how reports are phrased), \emph{scanner and site bias}, and
errors or shortcuts in automated preprocessing (e.g., LLM-based snippet extraction). Confident but incorrect
associations between images and text could contribute to
misdiagnosis, over-investigation, or inappropriate automation if
integrated into clinical workflows without calibration, uncertainty
communication, and governance. Deployment also imposes \emph{privacy}
obligations: training on patient data must follow institutional
review, data-use agreements, and de-identification standards appropriate
to each jurisdiction.
We therefore position \name{} as a methodological building block,
not a turnkey clinical product. We encourage translation efforts to
prioritize prospective evaluation, fairness and robustness auditing
across sites and demographics, clinician oversight, and transparent
reporting of limitations, including failure modes under distribution
shift and imperfect text or segmentation supervision.